%% file: main.tex
\documentclass{article}

\PassOptionsToPackage{numbers, compress}{natbib}
\usepackage[preprint,nonatbib]{neurips_2024}

\usepackage[utf8]{inputenc} 
\usepackage[T1]{fontenc}    
\usepackage{hyperref}       
\usepackage{url}            
\usepackage{booktabs}       
\usepackage{amsfonts}       
\usepackage{nicefrac}       
\usepackage{microtype}      
\usepackage{xcolor}         
\usepackage{amsmath}
\usepackage{amssymb}
\usepackage{enumitem}
\usepackage{graphicx}
\usepackage{wrapfig}
\usepackage{listings}
\usepackage{tabularx}
\usepackage{subcaption}

\definecolor{myred}{HTML}{FF8577}
\definecolor{mygreen}{HTML}{0FA958}
\definecolor{myblue}{HTML}{1982C4}
\definecolor{codegreen}{rgb}{0,0.5,0}
\definecolor{codegray}{rgb}{0.5,0.5,0.5}
\definecolor{codepurple}{rgb}{0.07,0,0.53}
\definecolor{codered}{RGB}{189,41,0}
\definecolor{codecomment}{RGB}{153,153,153}
\definecolor{backcolour}{rgb}{0.96,0.96,0.96}
\definecolor{royalblue}{rgb}{0.0, 0.14, 0.4}
\definecolor{egyptianblue}{rgb}{0.06, 0.2, 0.65}
\definecolor{royalazure}{rgb}{0.0, 0.22, 0.66}
\definecolor{portlandorange}{rgb}{1.0, 0.35, 0.21}
\definecolor{saddlebrown}{RGB}{139,69,19}
\definecolor{sienna}{RGB}{183,105,68}
\definecolor{saddlebrown}{RGB}{139,69,19}

\hypersetup{
    colorlinks=true,
    linkcolor=sienna,
    citecolor=egyptianblue,
}

\newtheorem{lemma}{Lemma}
\newtheorem{assumption}{Assumption}

\newcommand{\ie}{\textit{i}.\textit{e}.}
\newcommand{\eg}{\textit{e}.\textit{g}.}
\newcommand{\method}{\textsc{LoGra}}
\newcommand{\software}{\textsc{Logix}}

\title{\textit{What is Your Data Worth to GPT?}\\LLM-Scale Data Valuation with Influence Functions}

\author{
  \small{Sang Keun Choe$^1$\hspace{-0.3mm}\thanks{Lead author: \href{mailto:sangkeuc@andrew.cmu.edu}{sangkeuc@andrew.cmu.edu}.$\;\,^\dagger$Main contributors.}  $\;\,$Hwijeen Ahn$^{1\dagger}$ Juhan Bae$^{2\dagger}$ Kewen Zhao$^{1\dagger}$}\\[0.1ex]
  \small{\textbf{Minsoo Kang$^{3}$ Youngseog Chung$^{1}$ Adithya Pratapa$^{1}$ Willie Neiswanger$^{4}$}}\\[0.1ex]
  \small{\textbf{Emma Strubell$^{1}$ Teruko Mitamura$^{1}$ Jeff Schneider$^{1}$ Eduard Hovy$^{1}$ Roger Grosse$^{2}$ Eric Xing$^{1,5}$}}\\[0.1ex]
  \small{$^1\,$Carnegie Mellon University\hspace{0.2cm} $^2\,$University of Toronto\hspace{0.2cm} $^3\,$Georgia Tech\hspace{0.2cm} $^4\,$USC\hspace{0.2cm} $^5\,$MBZUAI}
}

\begin{document}

\maketitle

\begin{abstract}
  Large language models (LLMs) are trained on a vast amount of human-written data, but data providers often remain uncredited. 
  In response to this issue, data valuation (or data attribution\footnote{Noting that the leave-one-out error~\cite{koh2017understanding}, a basis for most data attribution methods, is a \textit{semivalue} \cite{dubey1981value,kwon2021beta,wang2023data}, we use ``data valuation'' as a unified term in this work.}), which quantifies the contribution or value of each data to the model output, has been discussed as a potential solution.
  Nevertheless, applying existing data valuation methods to recent LLMs and their vast training datasets has been largely limited by prohibitive compute and memory costs.
  In this work, we focus on influence functions, a popular gradient-based data valuation method, and significantly improve its scalability with an efficient gradient projection strategy called \method\ that leverages the gradient structure in backpropagation.
  We then provide a theoretical motivation of gradient projection approaches to influence functions to promote trust in the data valuation process.
  Lastly, we lower the barrier to implementing data valuation systems by introducing \software, a software package that can transform existing training code into data valuation code with minimal effort.
  In our data valuation experiments, \method\ achieves competitive accuracy against more expensive baselines while showing up to 6,500$\times$ improvement in throughput and 5$\times$ reduction in GPU memory usage when applied to Llama3-8B-Instruct and the 1B-token dataset (open source project: \href{https://github.com/logix-project/logix}{link}).

\end{abstract}

\input{introduction}
\input{background}
\input{method}
\input{experiments}
\input{related}

\input{conclusion}
\input{acknowledgement}

\bibliographystyle{plain}
\bibliography{reference.bib}

\input{appendix}

\end{document}

%% file: introduction.tex
\section{Introduction}
Despite the well-recognized importance of training data in advancing the capabilities of large language models (LLMs)~\cite{brown2020language,kaplan2020scaling,Razeghi2022ImpactOP}, there is no agreed-upon mechanisms for crediting or compensating data providers. As LLMs are increasingly integrated into our society and economy, the absence of such mechanisms has aggravated a tension between data and model providers, exemplified by recent legal challenges involving major tech companies~\cite{jlversusalphabet,metz2022lawsuit}. In this atmosphere, data valuation, which quantifies the contribution of each training data to the model output, has been discussed as a potential technical solution for tackling these societal issues~\cite{fernandez2023data,ghorbani2019data,huang2023citation,jia2019towards,worledge2023unifying,zhao2023addressing}. 

At a high level, most data valuation algorithms interpret the model output as a coalition of its training data, and evaluate the contribution of each example based on its influence on the model output when included or excluded from the training dataset~\cite{ghorbani2019data,ilyas2022datamodels,koh2017understanding,kwon2021beta}. If an inclusion of a specific training example consistently improves model performance, high value can be assigned to this example for its contribution. However, applying existing data valuation methods to recent LLMs and their vast training datasets has faced significant scalability challenges to date. For instance, sampling-based methods, such as the Shapley value~\cite{ghorbani2019data,kwon2021beta} or Datamodels~\cite{ilyas2022datamodels}, require retraining the model multiple times with varied combinations of data subsets to directly model the effect of in/excluding each data. Unfortunately, such repeated retraining is hardly affordable even for small models, let alone LLMs. To overcome this issue, gradient-based methods, including influence functions~\cite{koh2017understanding,park2023trak}, approximate the effect of data in/exclusion on the model output using gradient information without costly retraining. Even so, scaling gradient-based methods to LLMs is hindered by prohibitive compute and memory costs originating in the high-dimensional nature of the gradient.

Consequently, the main objective of this work is to bridge the gap in scaling existing data valuation methods to recent LLMs and their vast training datasets. Toward this goal, we focus on influence functions \cite{koh2017understanding,park2023trak}, a representative gradient-based data valuation method, and significantly improve its scalability with an efficient gradient projection algorithm. We visualize the proposed data valuation system in Figure~\ref{fig:diagram}, and detail our technical contributions below:

\begin{figure}
    \centering
    \includegraphics[width=0.94\textwidth]{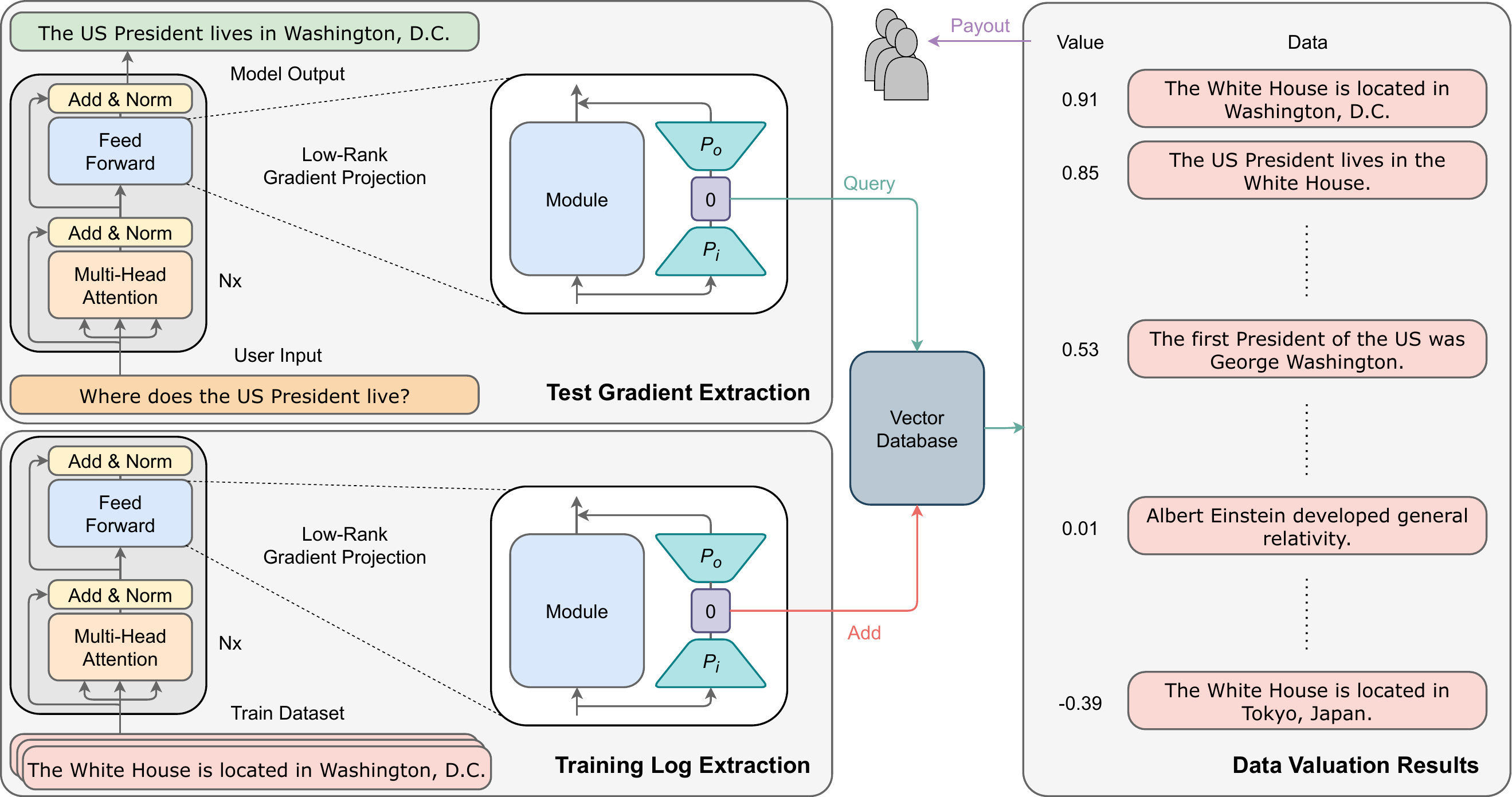}
    \vskip -4pt
    \caption{Data valuation system architecture. \textbf{(Left Bottom)} We first extract the Hessian and gradients for all training data using efficient gradient projection \method\ and store them in a database. \textbf{(Left Top)} At test time, we similarly extract gradients and query the database. \textbf{(Right)} The database returns similarity scores with respect to training examples that can be used for data valuation/attribution.}
    \label{fig:diagram}
\end{figure}

\begin{itemize}[leftmargin=*,topsep=-2pt]
    \item Employing gradient structures in backpropagation, we develop a novel \textbf{lo}w-rank \textbf{gra}dient projection algorithm \method\ that improves space \& time complexity of gradient projection, a major scalability bottleneck in prior work~\cite{park2023trak,schioppa2022scaling}, from $O(nk)$ to $O(\sqrt{nk})$ where $n$ and $k$ are model and projection dimensions. Furthermore, \method\ directly computes projected gradients without materializing full gradients, enabling low GPU memory and high GPU utilization for improved efficiency. Lastly, we show that \method\ can be easily implemented with small add-on layers, similarly to LoRA~\cite{hu2021lora}.
    \item By interpreting a damping term in influence functions as a spectral gradient sparsification mechanism, we (1) offer a theoretical motivation of gradient projection approaches to influence functions and (2) derive a specialized PCA initialization scheme for \method.
    \item We introduce software named \software\ that (1) makes it \textit{simple} to convert existing training code into data valuation code, (2) is \textit{compatible} with various scalability tools and features in the LLM ecosystem, and (3) is \textit{extensible} to implement other data valuation or interpretability algorithms.
    \item In our data valuation experiments, \method\ demonstrates competitive accuracy against more costly baselines, while showing up to 6,500$\times$ increase in throughput and 5$\times$ reduction in GPU memory, when applied to Llama3-8B-Instruct~\cite{llama3modelcard} and the 1B-token dataset, compared to EKFAC influence \cite{grosse2023studying}, the state-of-the-art and only runnable baseline at this scale. We also observe that most valuable data identified by \method\ generally share qualitative similarities with the queried LLM output.
\end{itemize}

%% file: background.tex
\section{Scalability Bottlenecks in Influence Functions}
\label{sec:background}
Most data valuation algorithms (\eg,\ data Shapley~\cite{ghorbani2019data}) evaluate the contribution or value of a specific example $x$ on the utility $v$ (\eg,\ test loss), that can further be used for crediting data providers, by measuring the overall change in the utility $v$ when in/excluding $x$ as follows:

\begin{align}
    \textsc{Value}(x;v) = \sum_{S\subseteq D\backslash \{x\}}w\big(v(S\cup\{x\}) - v(S)\big)\label{eq:shapley}
\end{align}
where $D$ is the training dataset,
$S$ is a subset of $D$, and $w$ is an (algorithm-specific) weighting term. Intuitively, the larger the utility gain from an inclusion of $x$ is, the larger the value of $x$ is.

One popular instantiation of Eq.~\eqref{eq:shapley} is the leave-one-out error~\cite{koh2017understanding}, a semivalue~\cite{dubey1981value} that is a basis for most data attribution methods and only considers $S$ with $|S|=|D|-1$ (\ie,\ leaving one example $x$ from the entire dataset $D$).
However, naively computing the leave-one-out-error requires retraining the model multiple times for each $x\in D$, which is hardly affordable even in small-scale setups.
To overcome this issue, influence functions, a representative gradient-based method, efficiently \textit{simulates} the effect of model retraining without an example $x_{tr}$ on the utility using gradient information as:
\begin{align}
    \textsc{Influence}(x_{tr}, x_{te}) = g_{te}^\top H^{-1}g_{tr}\label{eq:if}
\end{align}
where $g_{tr}$ and $g_{te}$ are train and test gradients respectively, and $H$ is the Hessian matrix. Concretely, influence functions approximate the effect of removing $x_{tr}$ by updating the model parameters with a Newton step in the direction of $H^{-1}g_{tr}$, and uses a first-order Taylor approximation to estimate how this update will affect the test utility. In practice, computing influence functions involves two key steps of (1) solving the inverse Hessian-vector product (iHVP) with $g_{te}$, and (2) taking the dot product of this iHVP with the gradient $g_{tr}$ for each training example.

Despite their comparative efficiency, influence functions remain difficult to scale to recent LLMs due to the high compute and memory costs associated with both steps.
First, space and time complexity of naive iHVP are respectively $O(n^2)$ and $O(n^3)$, both of which are impractical in recent LLMs with $n>10^9$ parameters. To address this issue, various tricks for efficiency, such as iterative methods \cite{koh2017understanding} or EKFAC approximation~\cite{grosse2023studying}, have been proposed. Second, to ensure fair valuation, one must compute influence scores with \textit{all} training data, which requires access to their gradients. However, computing gradients for all training data approximately amounts to one-epoch training, the cost of which often exceeds \$1M in the context of LLM (pre)training. If training gradients were to be recomputed frequently for regular data valuation, the total cost can quickly become astronomical. Thus, while it is technically possible to run a few influence function analyses to interpret interesting LLM outputs using efficient iHVP tricks~\cite{grosse2023studying}, doing it in a scalable and sustainable way to build a practical data valuation system remains a significant challenge.

In an attempt to mitigate the aforementioned cost issues, Arnoldi IF~\cite{schioppa2022scaling} and TRAK~\cite{park2023trak} recently explored the strategy of projecting gradients onto a low-dimensional space and computing influence scores on the subspace spanned by the projection matrix as follows:
\begin{align}
    \textsc{Influence}(x_{tr}, x_{te};P) =\big(Pg_{te}\big)^\top\big(PHP^\top\big)^{-1}\big(Pg_{tr}\big)\label{eq:if_proj}
\end{align}
where $P\in\mathbb{R}^{k\times n}$ is the projection matrix given the model and projection dimensions of $n$ and $k$. Under this strategy, the iHVP operation also occurs in a low-dimensional space, meaning that $n$ in memory and compute complexity of iHVP gets replaced with $k\ll n$. Furthermore, low-rank projection enables writing projected gradients for all training data to disk once and simply reading them as new test data arrives without costly re-computations. This converts an influence function problem  into a vector similarity search problem, for which various system optimizations exist~\cite{johnson2019billion}.

In essence, this strategy significantly reduces both iHVP and training gradient recomputation costs by introducing an additional process of low-rank gradient projection $Pg$. However, the additional compute/memory costs and accuracy degradation incurred from low-rank gradient projection has not been thoroughly studied to date. First, assuming that the batch size is $b$, the compute cost of naive batched gradient projection is $O(bkn)$. Noting that the compute cost of backpropgation is $O(bn)$ (or $O(btn)$ if we consider the time dimension), the cost of gradient projection is usually larger than that of backpropagation given a reasonably large $k$ for the expressivity. Second, the memory costs for full per-sample gradient and the projection matrix are $O(bn)$ and $O(kn)$. If an 8B model were to be used, each of these costs amounts to 32GB$\times b$ (or $\times k$) GPU memory. While Arnoldi IF and TRAK attempt to address the memory costs of the per-sample gradient and projection matrix respectively with forward-mode Jacobian-vector products and a custom CUDA kernel trick, neither of them are able to solve both issues altogether. This leads Arnoldi IF and TRAK to use very small $k$ and $b$, each of which results in decreased accuracy of influence scores due to limited expressivity and poor efficiency from low GPU utilization. Since accuracy and efficiency are both critical for effective data valuation, we deduce that further advancements in the gradient projection approach are necessary.

%% file: method.tex
\section{Scaling Data Valuation \& Influence Functions}
\label{sec:method}
In light of these issues, we first design a memory and compute efficient gradient projection algorithm called \method, that leverages the inherent gradient structure in backpropagation (Section~\ref{sec:algorithm}). Then, we provide an intuitive theoretical analysis on why gradient projection approaches work in influence functions (Section~\ref{sec:theory}). Finally, we distill our insights obtained from studying (scalable) influence functions into a new open-source software, called \software, which achieves high compatibility, extensibility, and usability, to facilitate data valuation research  (Section~\ref{sec:software}). In this section, we build our arguments at the granularity of each layer (or module) instead of the whole network for clarity.

\subsection{Algorithm: Memory and Compute Efficient Gradient Projection}
\label{sec:algorithm}
Most layers in neural networks, such as linear and convolutional layers, essentially perform matrix multiplication. Given the input $x_i\in\mathbb{R}^{n_i\times T}$, the output $x_o\in\mathbb{R}^{n_o\times T}$, the weight $W\in\mathbb{R}^{n_o\times n_i}$ for the layer, its forward and backward computations can be written as follows:
\begin{flalign}
\text{\textbf{Forward:}}&&x_o =&\,Wx_i&&\\[-0.1ex]
\text{\textbf{Backward:}}&&\text{vec}(\mathcal{D}W)=\sum_{t=1}^Tx_{i,t}\,\otimes&\, \mathcal{D}x_{o,t}\,,\;\;\mathcal{D}x_i=W^\top\mathcal{D}x_o&&\label{eq:backprop}
\end{flalign}
where $T$ denotes for the sequence dimension in language modeling, $\mathcal{D}$ the derivative with respect to the loss, $\otimes$ the Kronecker product, and $\text{vec}(\cdot)$ the vectorization operation. In Eq.~\eqref{eq:backprop}, we observe that gradient $\text{vec}(\mathcal{D}W)$ obtained during backpropagation is structured as a sum of Kronecker products between forward and backward activations. \method\ leverages this observation to impose an additional Kronecker-product structure on the projection matrix $P$ as follows:
\begin{align}
    P\text{vec}(\mathcal{D}W)\triangleq(P_i\otimes P_o)\text{vec}(\mathcal{D}W)=\sum_{t=1}^T(P_i\otimes P_o)(x_{i,t}\otimes\mathcal{D}x_{o,t})=\sum_{t=1}^T P_ix_{i,t}\otimes P_o\mathcal{D}x_{o,t}\label{eq:logra}
\end{align}
where $P_i\in\mathbb{R}^{k_i\times n_i}$, $P_o\in\mathbb{R}^{k_o\times n_o}$, and $P=P_i\otimes P_o$. In Eq.~\eqref{eq:logra}, \method\ first projects forward and backward activations onto low-dimensional spaces with $P_i$ and $P_o$ respectively, and then reconstructs projected gradient directly from these projected activations. This is in contrast to traditional gradient projection~\cite{park2023trak}, which first computes raw gradient and then projects it onto a low-dimensional space.

Now, we compare memory/compute efficiency of \method\ to that of naive gradient projection, especially under the setting of $n_i\approx n_o\approx\sqrt{n}$ and $k_i\approx k_o\approx\sqrt{k}$. First, both memory/compute costs of per-sample gradient computations reduce from $O(bn)$ to $O(bk)$. Second, both memory/compute costs of gradient projection reduce from $O(bnk)$ to $O(b\sqrt{nk})$. To clearly see this benefit, given the model/projection sizes of 8B/4k, we note that projection matrix sizes are about 1GB and 128TB respectively for \method\ and naive projection. As such, while enjoying general efficiency gains from gradient projection we disscussed in Section~\ref{sec:background}, \method\ further improves the efficiency of per-sample gradient computations significantly at a marginal cost of the additional gradient projection process.

\begin{wrapfigure}{r}{0.26\textwidth}
  \begin{center}
  \vskip -18pt
    \includegraphics[width=0.26\textwidth]{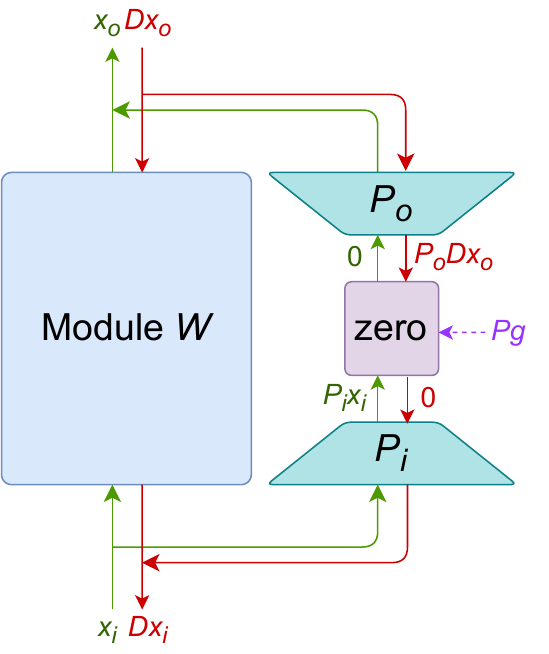}
  \end{center}
  \vskip -8pt
  \caption{\method.}
  \vskip -10pt
  \label{fig:logra}
\end{wrapfigure}

Furthermore, leveraging the fact that projection occurs in the activation space, \method\ can be easily implemented with small add-on layers that are composed of \textit{encoder}, \textit{bottleneck}, and \textit{decoder}, each of which is initialized with $P_i$, zero, and $P_o$ as shown in Figure~\ref{fig:logra}. If we ignore the bottleneck layer, the overall architecture is identical to the popular LoRA architecture \cite{hu2021lora}. While it is intuitive that the roles of encoder and decoder are projecting forward and backward activations respectively, we emphasize two critical roles of the bottleneck layer here. First, its zero initialization ensures that the rest of both forward and backward computations remain unaffected by these add-on layers. Second, per-sample projected gradients can be obtained by simply computing per-sample gradients for the bottleneck layer, using automatic differentiation of an underlying framework without complicated implementation efforts.

\subsection{Theory: Why Gradient Projection Works in Influence Functions}
\label{sec:theory}
While \method\ can significantly improve scalability of influence functions, an inherent criticism of any gradient projection approach is that information loss from the projection process may render the resulting influence analysis invalid. Unfortunately, theoretical analyses from prior work~\cite{park2023trak,schioppa2022scaling} only discuss the indirect effect of gradient projection on proxy concepts like gradient flow or iHVP variance, which are loosely related to influence functions. To promote trust in the data valuation process, we provide here a mathematical motivation of gradient projection approaches to influence functions. Toward this goal, we interpret a damping term in influence functions that is typically added to ensure the invertibility of the Hessian $H$ as a \textit{spectral gradient sparsification} mechanism. A formal argument and our derivation are respectively provided in Lemma~\ref{eq:lemma} and in Appendix~\ref{sec:derivation}.
\begin{lemma}
\label{eq:lemma}
Let $\{e_1,\cdots,e_n\}$ and $\{\lambda_1,\cdots,\lambda_n\}$ be eigenvectors and eigenvalues of the Hessian $H$. Expressing $g_{tr/te} =\sum_ic_{tr/te,i}\cdot(\sqrt{\lambda_i}e_i)$, the following holds under Assumption~\ref{eq:assumption1}:
\begin{align*}
  \textsc{Influence}(x_{tr}, x_{te}) = g_{te}^\top (H+\lambda I)^{-1}g_{tr} = \sum_{i=1}^n\frac{\lambda_i}{\lambda_i+\lambda}c_{tr,i}c_{te,i}\;\;\text{and}\;\;\mathbb{E}[c_{\cdot,i}^2]\approx 1.  
\end{align*}
\end{lemma}
Lemma~\ref{eq:lemma} shows that a damping term \textit{softly} limits the number of components in influence computations by penalizing contributions from small components. Given the prevalence and practical importance of a damping term in influence functions~\cite{basu2020influence}, we can motivate gradient projection as an alternative way of (hard-)limiting influence computations to components in the projection matrix.
To make \method\ similarly penalize small components, we develop an initialization scheme that exploits the Kronecker-Factored Approximate Curvature (KFAC) algorithm~\cite{martens2015optimizing}.
As a quick overview, KFAC approximates the block-wise Hessian with the Kronecker product of uncentered forward and backward covariances of each layer, respectively denoted with $C_F$ and $C_B$, as $H\approx H_{KFAC} = C_F \otimes C_B$.
Expressing $C_F$ and $C_B$ as $Q_F\Lambda_FQ_F^\top$ and $Q_B\Lambda_BQ_B^\top$ with eigendecomposition, it is easy to show that eigenvectors and eigenvalues of $H_{KFAC}$ are $Q_F\otimes Q_B$ and $\Lambda_F\otimes \Lambda_B$. Consequently, we can approximately discard the smaller components of $H$ by initializing $P_i$ and $P_o$ with $Q_F^{1:k_i}$ and $Q_B^{1:k_o}$, where $Q_\cdot^{1:k}$ is a collection of top-$k$ eigenvectors (similar to performing PCA on forward and backward activations). In Section~\ref{sec:experiments}, we experiment with both PCA and random initialization schemes.

\subsection{Software: Compatibility, Extensibility, and Usability}
\label{sec:software}
Besides algorithmic efficiency, another major bottleneck in the practical adoption of data valuation systems is often the challenge of implementation. In particular, we observe that gradient computation in LLMs, which is a building block for influence functions, typically requires support from other scalability tools like DeepSpeed~\cite{rasley2020deepspeed} or relies on high-level frameworks like HF Transformers~\cite{wolf-etal-2020-transformers}. However, most existing software that can be used for data valuation (\eg,\ Captum~\cite{kokhlikyan2020captum} and TRAK \cite{park2023trak}) is largely incompatible with these tools due to the (too) high level of abstraction in their APIs. 

\begin{wrapfigure}{r}{0.47\textwidth}
\vskip -18pt
\begin{lstlisting}
import logix

# setup
run = logix.init(project, config)
run.setup("stat": "kfac", "save": "grad")
run.watch(model)

# train log & statistic
for batch in train_loader:
  with run(data_id=batch["input_ids"]):
    loss = model(batch)
    loss.backward()
run.finalize()

# test time influence analysis
with run(data_id=tst_batch["input_ids"]):
  loss = model(tst_batch)
  loss.backward()
run.compute_influence_all()
\end{lstlisting}
\vskip -10pt
\caption{Code Example of \software.}
\vskip -11pt
\label{fig:code}
\end{wrapfigure}
Subsequently, we develop a new software package, \software, design of which enables an \textit{easy conversion} of users' existing training code into data valuation code, by promoting compatibility with other tools in the LLM ecosystem.
To this end, we first notice that most influence function algorithms simply require collecting train logs (\eg,\ gradient, activation) and their statistics (\eg,\ covariance). As a result, given arbitrary users' training code, data valuation software only need to intercept these logs, and provide basic primitives to compute various statistics with them. Leveraging this observation, \software\ implements log interceptions and compute primitives using PyTorch hooks. Notably, the use of hooks makes \software\ \textit{compatible} with diverse other tools as hooks can be seamlessly integrated with most PyTorch features (\eg,\ FSDP, autocast, compile). In addition, \software\ is \textit{extensible}, as users can easily define and add custom primitives inside hooks. Finally, \software\ is \textit{easy-to-use} as its context manager automatically handles adding appropriate hooks and primitives to relevant modules with minimal code changes. In Appendix~\ref{sec:logix_appendix}, we provide a more detailed comparison between \software\ and other relevant (interpretability) software, and describe notable optimization techniques (\eg,\ efficient data IO) implemented in it.
Code examples can be found in Figure~\ref{fig:code}, Appendix~\ref{sec:code}, and our project \href{https://github.com/logix-project/logix}{page}.

%% file: experiments.tex
\section{Experiments}
\label{sec:experiments}
In this section, we evaluate the effectiveness of \method\ in terms of \textit{accuracy} and \textit{efficiency}, both of which are important in practical data valuation systems. Specifically, we first perform two types of counterfactual evaluations to quantitatively study data valuation accuracy of \method\ on small-scale setups (Section~\ref{sec:brittleness}). Then, we scale \method\ to LLMs and their massive training data, where we investigate qualitative accuracy (\ie,\ how similar most valuable training data are to the model output) and memory/compute efficiency (Section~\ref{sec:llm}). Finally, our appendix includes more qualitative results of data valuation (Appendix~\ref{sec:qualitative}), pseudo-code for LLM experiments (Appendix~\ref{sec:code}), and experimental details such as hyperperameters and compute resources (Appendix~\ref{sec:hyperparams}).

\subsection{Quantitative Accuracy with Counterfactual Evaluation}
\label{sec:brittleness}

\begin{figure}[htbp]
    \centering
    \begin{subfigure}[b]{\textwidth}
        \centering
        \includegraphics[width=0.97\textwidth]{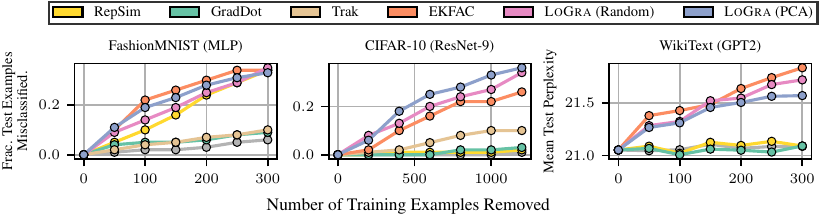}
        \vskip -3pt
        \caption{Brittleness test}
        \label{fig:subset}
    \end{subfigure}
    \vskip 5pt
    \begin{subfigure}[b]{\textwidth}
        \centering
        \includegraphics[width=0.97\textwidth]
        {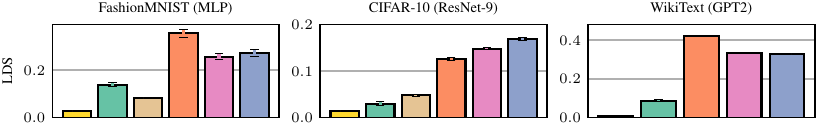}
        \vskip -3pt
        \caption{Linear datamodeling score (LDS)}
        \label{fig:lds}
    \end{subfigure}
    \vskip -3pt
    \caption{Quantitative accuracy evaluation of data valuation algorithms. We excluded TRAK in the WikiText experiments due to lack of a public implementation for language modeling tasks.}
    \label{fig:quantitative}
\end{figure}

To quantitatively assess accuracy of data valuation algorithms, we adopt two counterfactual evaluation methods: brittleness test~\cite{ilyas2022datamodels} and linear datamodeling score (LDS)~\cite{park2023trak}. First, the brittleness test focuses on accuracy in successfully identifying \textit{top} valuable data. To this end, it first removes the top-$k$ valuable data identified by each algorithm, retrains the model without them multiple times with different random seeds, and measures the overall change in the model output. The larger the output change is, the more accurate the algorithm is in identifying \textit{top} valuable data. Second, LDS measures general valuation accuracy of \textit{all} training data under the additivity assumption. Specifically, given multiple data subsets $\{S_i\}$ of the fixed size (\eg,\ $|S_i|=|D|/2$), LDS estimates the test performance of the model trained on $S_i$ by summing the values of all examples in $S_i$ returned by each algorithm, and compares it against the gold performance obtained by actually training the model on $S_i$ using the Spearman correlation. Noting that linear datamodels have a connection to the game-theoretic data value (\eg,\ Shapley value)~\cite{ilyas2022datamodels}, LDS can serve as a principled way to study data valuation accuracy.

We perform these counterfactual evaluations on three benchmarks where many rounds (up to 1800) of retraining is feasible: (1) MLP with FMNIST, (2) ResNet-9~\cite{he2016deep} with CIFAR-10, and (3) GPT2~\cite{radford2019language} with WikiText. On these benchmarks, we compare accuracy of \method\ against four popular data valuation baselines, including gradient dot product~\cite{pruthi2020estimating}, TRAK~\cite{park2023trak}, EKFAC influence~\cite{grosse2023studying}, and representation similarity~\cite{hanawa2020evaluation}. With the aim of bearing relevance to a large-scale setting with LLMs and their vast training data, we have only considered baseline methods that satisfy the following two conditions. First, the method cannot retrain the model multiple times for identifying top-$k$ valuable data.\footnote{Note that multiple retraining is only allowed for evaluating accuracy of already identified top-$k$ data, but not for identifying top-$k$ data itself in our experiments.} Second, the method only has access to the final model checkpoint, which is the case for most LLMs. Given the above setup, we present our experiment results in Figure~\ref{fig:quantitative}.

We observe that \method\ slightly underperforms EKFAC influence, which is a few orders of magnitude slower in our large-scale experiments (Section~\ref{sec:llm}), while noticeably outperforming other baselines. We attribute competitive accuracy of \method\ to two factors. First, unlike TRAK of which projection dimension is limited by the huge projection matrix, \method\ can efficiently afford a higher projection dimension thanks to its sublinear memory/compute costs for gradient projection, and thus achieve the higher expressivity.
Second, gradient projection enables \method\ to compute raw projected Fisher information matrix (or Hessian) without an approximation as in EKFAC influence. We expect that a more accurate computation of the Hessian generally leads to more accurate data valuation results.

Comparing the initialization schemes for \method\ (PCA vs. random), we observe that \method-PCA outperforms \method-random on the FMNIST and CIFAR benchmarks. Hence, we hypothesize that it is generally more accurate to compute influence functions with larger components, similar to the spectral gradient sparsification effect of a damping term we discussed in Section~\ref{sec:theory}. To understand a relatively poor performance of \method-PCA on WikiText+GPT2, we point out that the Transformer architecture~\cite{vaswani2017attention} used in this benchmark lacks the specialized KFAC Hessian approximation, unlike naive MLP~\cite{martens2015optimizing} or convolutional~\cite{grosse2016kronecker} architectures in other benchmarks. Subsequently, our ad-hoc implementation of the PCA initialization based on the naive MLP architecture (\ie,\ no weight sharing) may not successfully keep larger components of the GPT2 Hessian, failing to deliver its benefit. As a result, we decide to use \method-random for our LLM experiments in the next subsection.

\subsection{Scaling to Billion-Scale Models \& Datasets}
\label{sec:llm}
Given competitive accuracy of \method, we now evaluate its practical utility in valuing billion-scale training data for billion-scale models. Specifically, we adopt GPT2-XL (1.5B) \cite{radford2019language}, Pythia-1.4B~\cite{biderman2023pythia}, and Llama3-8B-Instruct~\cite{llama3modelcard} as our models, and conduct data valuation on a random 1B-token subset of the OpenWebText (OWT) dataset~\cite{gokaslan2019openwebtext}. The major motivations behind choosing OWT as our data valuation dataset are twofold. First, we observe that OWT consists of relatively higher-quality data compared to other LLM training datasets like C4~\cite{raffel2020exploring} or Dolma~\cite{soldaini2024dolma} while maintaining the diversity unlike other high-quality datasets like WikiText~\cite{merity2016pointer}. Second, we anticipate that OWT largely overlaps with training datasets of all our models. In detail, GPT2-XL is trained on the WebText dataset that shares the same data curation process with OWT, Pythia-1.4B is trained on the Pile dataset~\cite{gao2020pile} that includes an extension of OWT (\ie,\ OpenWebText2), and we suppose a majority of OWT would be a part of Llama3's massive 15T-token pretraining dataset. We also note that our OWT subset size (\ie,\ 1B tokens) was mainly limited by the available storage, not by compute (see Table~\ref{tab:llm_efficiency}). If we had access to a storage size of 1PB, performing data valuation with a dataset size of 100B+ tokens would be readily feasible using the same compute resource.

\textbf{Efficiency.\hspace{2.5mm}} To begin with, we compare memory and compute efficiency of \method\ against EKFAC influence~\cite{grosse2023studying}, the state-of-the-art and only algorithm that can run on billion-scale models without CUDA out-of-memory (OOM) errors. Indeed, we confirm that running TRAK or Arnoldi IF with billion-scale models results in CUDA OOM errors even on A100 GPUs with 80GB VRAM due to their gigantic projection matrix sizes. We report GPU memory usage and throughput of both logging (one-time) and influence computation (recurring) phases for the Llama3-8B-Instruct experiment with one A100 GPU and half-precision in Table~\ref{tab:llm_efficiency}.
\vskip -5pt
\begin{table}[htbp]
\centering
\scriptsize
\begin{tabularx}{\textwidth}{X*{8}{c}}
\toprule
       & \multicolumn{4}{c}{\textbf{Logging} (Compute \& save Hessian$\,$|$\,$grad)} & \multicolumn{4}{c}{\textbf{Compute Influence} (Dot product between test \& train grads)}  \\\cmidrule(r{4pt}){2-5} \cmidrule(l){6-9}
       & $\;$Batch$\;$ & $\;$Throughput$\;$     & $\;$Memory$\;$   & $\;$Storage$\;$ & $\;$Train Batch$\;$ & $\;$Test Batch$\;$ & $\;$Throughput$\;$  & $\;$Memory$\;$\\\midrule
EKFAC &1 & 1740 / 419$^*$          & 71 / 80$^*$GB & \textbf{89 GB}     & 4           & 4          & 12.2      & 75 GB  \\[0.1ex]
\method &1 & 3430 & \textbf{23 GB}   & 3.5 TB   & 256           & 4          & 1599.6      & \textbf{14 GB}  \\[0.1ex]
\method &16 & \textbf{4696} & 79 GB  & 3.5 TB    & 256           & 256          & \textbf{79003.9}      & \textbf{15 GB} \\\bottomrule
\end{tabularx}
\vskip 3pt
\caption{Memory \& compute efficiency analyses for \method\ and EKFAC. Throughput is measured as tokens/s for logging and (train, test) pairs/s for influence computations. $^*$ EKFAC logging consists of two subphases of KFAC fitting (left of /) and corrected eigenvalue fitting (right of /).}
\label{tab:llm_efficiency}
\end{table}

Due to the huge size of raw gradients (\eg,\ 16GB in fp16 for an 8B model), EKFAC cannot afford storing raw gradients for \textit{all} training data to disk.
As a result, EKFAC needs to recompute all training gradients for each test batch, and thus requires allocating extra GPU memory on model weights and intermediate activations. This largely limits both train/test batch sizes and throughput (12.2 pairs/s), and performing data valuation with EKFAC for 256 test data and 1B-token training data would take 11,300 A100 GPU hours, rendering it hardly usable in most practical setups.

In contrast, with its (efficient) gradient projection, \method\ not only significantly improves compute and memory efficiency, but also avoids training gradient recomputations at the costs of disk space for storing \textit{projected} training gradients and latency from data IO. Since the storage cost is typically much cheaper than the compute cost\footnote{\eg,\ hourly rates for a 1TB storage and one A100 GPU are approximately \$0.03 and \$4 on AWS.}, we believe our trade-off offers considerable practical benefits. Furthermore, we can largely hide the data IO cost by overlapping gradient reading/writing processes with other computations. For instance, given the fixed train gradient batch size of 256 (\ie,\ fixed data loading time), we are able to successfully overlap the process of loading training gradients from disk with influence computations against up to 256 test gradients, and thereby achieve almost 6,500$\times$ improvement in throughput from EKFAC influence. Noting that our GPU memory usage is far from saturated even with the train/test batch size of 256, we believe that more throughput improvements can be achieved simply by further increasing train/test batch sizes.

\textbf{Qualitative Accuracy.\hspace{2.5mm}} Next, we analyze qualitative similarities between queried LLM outputs and most valuable data identified by \method\ that can be critical for promoting trust in the data valuation system~\cite{worledge2023unifying}. Importantly, we observe that naive influence functions frequently return outlier data with high gradient norms as most valuable data, as also noted in \cite{barshan2020relatif,grosse2023studying}. To mitigate this issue, we instead use $l$-RelatIF, a variant of influence functions that normalizes the original influence score with the self-influence score of each training data to penalize such outlier effects~\cite{barshan2020relatif}. Our experimental results are provided in Figure~\ref{fig:llm} (concise) and in Appendix~\ref{sec:qualitative} (extensive).

\begin{figure}
    \centering
    \begin{subfigure}[b]{\textwidth}
        \centering
        \includegraphics[width=0.98\textwidth]{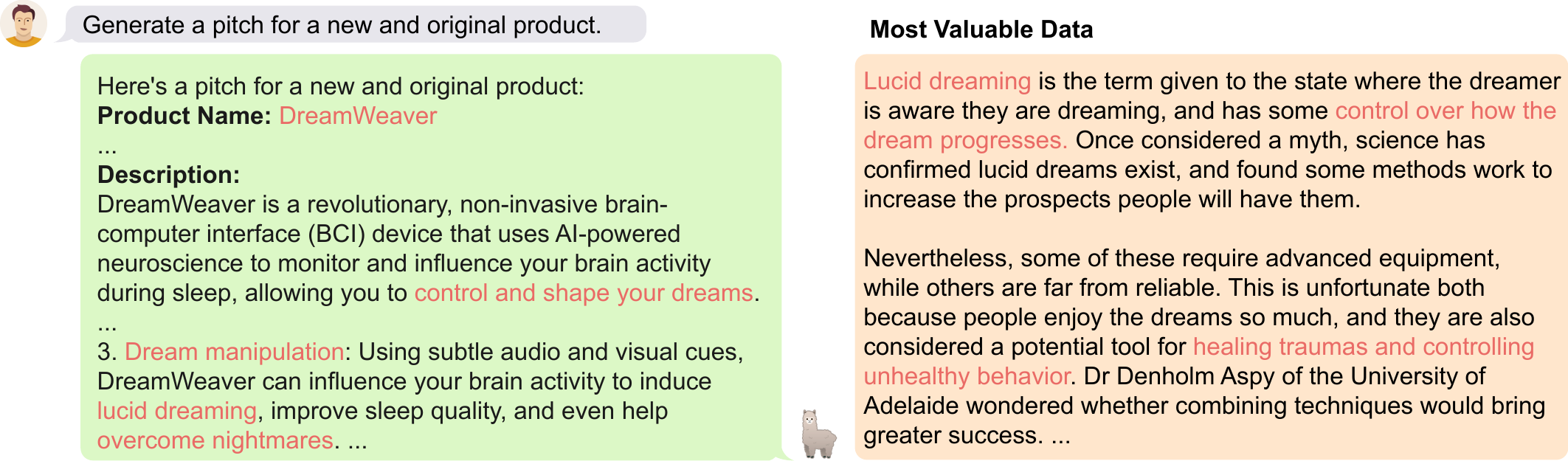}
        \vskip -3pt
        \caption{Llama3-8B-Instruct}
        \label{fig:llama3}
    \end{subfigure}

    \begin{subfigure}[b]{\textwidth}
        \centering
        \includegraphics[width=0.98\textwidth]{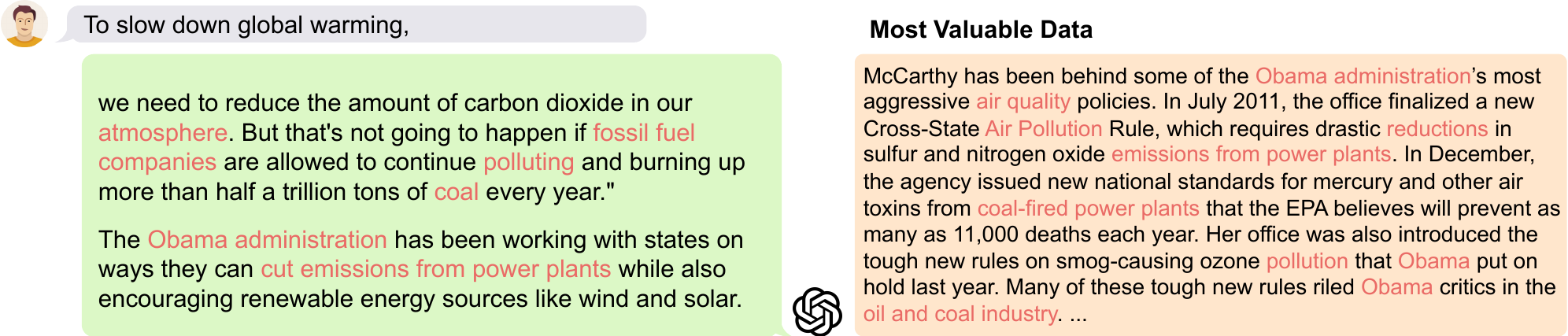}
        \vskip -3pt
        \caption{GPT2-XL (1.5B)}
        \label{fig:gpt2}
    \end{subfigure}

    \begin{subfigure}[b]{\textwidth}
        \centering
        \includegraphics[width=0.98\textwidth]{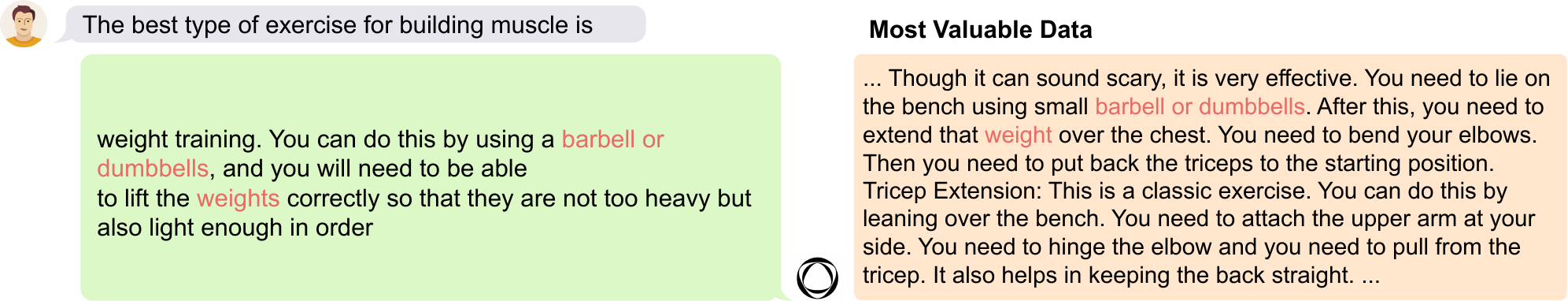}
        \vskip -3pt
        \caption{Pythia-1.4B}
        \label{fig:pythia}
    \end{subfigure}

    \caption{Qualitative accuracy of data valuations with \method. Important keywords in each example are \textit{manually} highlighted for the improved readability. More examples can be found in Appendix~\ref{sec:qualitative}.}
    \label{fig:llm}
\end{figure}

We observe that most valuable data identified by \method, especially for Llama3-8B-Instruct and GPT2-XL, share qualitative similarities (\eg,\ semantics, style, token overlaps) with the queried LLM outputs. For instance, given Llama3's response on the dream manipulation product, \method\ identifies a scientific article that studies manually inducing the lucid dream as most valuable data in Figure~\ref{fig:llama3}. In Figure~\ref{fig:gpt2}, both the GPT2-XL output and the corresponding most valuable data discuss the need for reducing emissions in the coal industry and its connection to the specific administration. In Figure~\ref{fig:pythia}, the concept of ``lifting barbell or dumbells'' appear in the model output and the most valuable data. 

However, we also notice several failure cases where the identified most valuable data seemingly do not share qualitative similarities with the LLM output, especially with Pythia-1.4B (Appendix~\ref{sec:pythia_appendix}). We here provide three potential explanations on these failing examples based on our experiments. First, attributed data may lack qualitative similarities when the queried LLM output itself is incoherent that its gradient does not encode meaningful information. This aligns with our observation that the failure case occurs more frequently with lower-tier models like Pythia-1.4B whose outputs generally are of lower quality. Second, since we only used a 1B-token subset for data valuation, it is possible that our valuation dataset may lack similar data to some queries. As noted above, our experiment was largely limited by the storage of our cluster (not by the compute), so exploring data valuation on an industry-scale cluster would be interesting future work. Third, we posit that train/test gradients in influence functions may encode diverse information including features that are hardly perceptible to humans~\cite{ilyas2019adversarial}. Therefore, it is possible that attributed data are indeed valuable for increasing the likelihood of the queried output by contributing to these other aspects while sharing little qualitative similarities. A more extensive argument on this final point can be found in Appendix~\ref{sec:pythia_appendix}.

%% file: related.tex
\section{Related Work}
\label{sec:related}
\textbf{Data Valuation.\hspace{2.5mm}} Measuring the value (or contribution) of training data on the model outputs has gained lots of attention recently. Exemplified by Data Shapley~\cite{ghorbani2019data}, a flurry of prior work~\cite{jia2019towards,kwon2021beta,wang2023data} proposed exploiting the Shapley value or concepts from cooperative game theory to address the data valuation problem. However, most existing approaches in this line require repeated retraining of the model, a cost of which is hardly affordable even with small models. In addition to game-theoretic approaches, data valuation has also been tackled using reinforcement learning~\cite{yoon2020data}, meta learning~\cite{choe2024making}, and training-free methods~\cite{nohyun2023data,wu2022davinz}. Nevertheless, these works either suffer from high complexity from the need to train other models~\cite{choe2024making,yoon2020data} or high computational costs~\cite{nohyun2023data}. We direct readers to Sim et al.~\cite{sim2022data} for a more extensive survey on diverse data valuation approaches.


\textbf{Influence Functions.\hspace{2.5mm}} Influence functions, a classic concept from robust statistics \cite{hampel1974influence}, estimate the infinitesimal effect of removing or adding a training data point without model retraining. They have various applications in machine learning, such as interpreting the model's behavior~\cite{han2020explaining,park2023trak,grosse2023studying} and curating training datasets~\cite{liu2021influence,engstrom2024dsdm}. However, when applied to large neural networks, the computation of the iHVP and its dot product with all training examples introduce scalability challenges. Besides gradient projection, past works have explored computing influence functions only on the last (few) layers~\cite{koh2017understanding,schioppa2022scaling} to mitigate these challenges. However, subsequent works~\cite{feldman2020neural,grosse2023studying} have shown that the influence on only a subset of layers is insufficient to capture the overall influence of a training data point. To avoid computing the gradient of all training examples, various filtering strategies, such as using the similarity in the model's representation space~\cite{guo2020fastif} or TF-IDF~\cite{yeh2022first,grosse2023studying}, have also been proposed. While it is possible to adopt these filtering strategies for \method, they may introduce bias in the selection of the most influential sequences. For example, filtering candidate training sequences with TF-IDF might miss interesting influential sequences that do not share many tokens but are semantically related. Recently, similarly to \method, DataInf~\cite{kwon2024datainf} and LESS~\cite{xia2024less} proposed using LoRA to efficiently compute influence functions. However, these approaches are only applicable in finetuning settings, whereas \method\ also supports influence analyses for pretraining.

%% file: conclusion.tex
\section{Conclusion}
In this work, we explored scaling data valuations with influence functions to billion-scale models and datasets as a potential technical solution to properly credit or compensate data providers for training LLMs. Toward this goal, we developed a novel gradient projection algorithm that can significantly improve the scalability of influence functions, and designed a simple and interoperable software. Our experiments showed that \method\ achieves competitive accuracy to other more expensive baselines on counterfactual evaluations, while efficiently scaling to billion-scale models and datasets, thereby demonstrating the initial potential of the practical data valuation system. Last but not least, we discuss broader impacts and limitations of our work in Appendix~\ref{sec:neurips}.

%% file: acknowledgement.tex
\section*{Acknowledgement}
We thank Daphne Ippolito, Shaily Bhatt, and Yongchan Kwon for providing insightful feedback in preparing the early version of the manuscript, and Jared Fernandez for the useful discussion on the LLM data valuation experiment. We acknowledge the CMU Babel cluster and its administrators for reliably providing necessary compute resources for this work.

%% file: appendix.tex
\newpage
\appendix

\section{Qualitative Analysis}
\label{sec:qualitative}
In this section, we provide more qualitative analyses on most valuable data identified by \method. In particular, we show top-2$\sim$4 valuable data for each query here, given the possibility that the model utilizes information from multiple training examples for generating its output. We also include several failure cases where identified most valuable data do not share qualitative similarities with the queried LLM output.

\subsection{Llama3-8B-Instruct}

\subsubsection{Example 1}
\begin{figure}[htbp]
    \centering
    \includegraphics[width=0.99\textwidth]{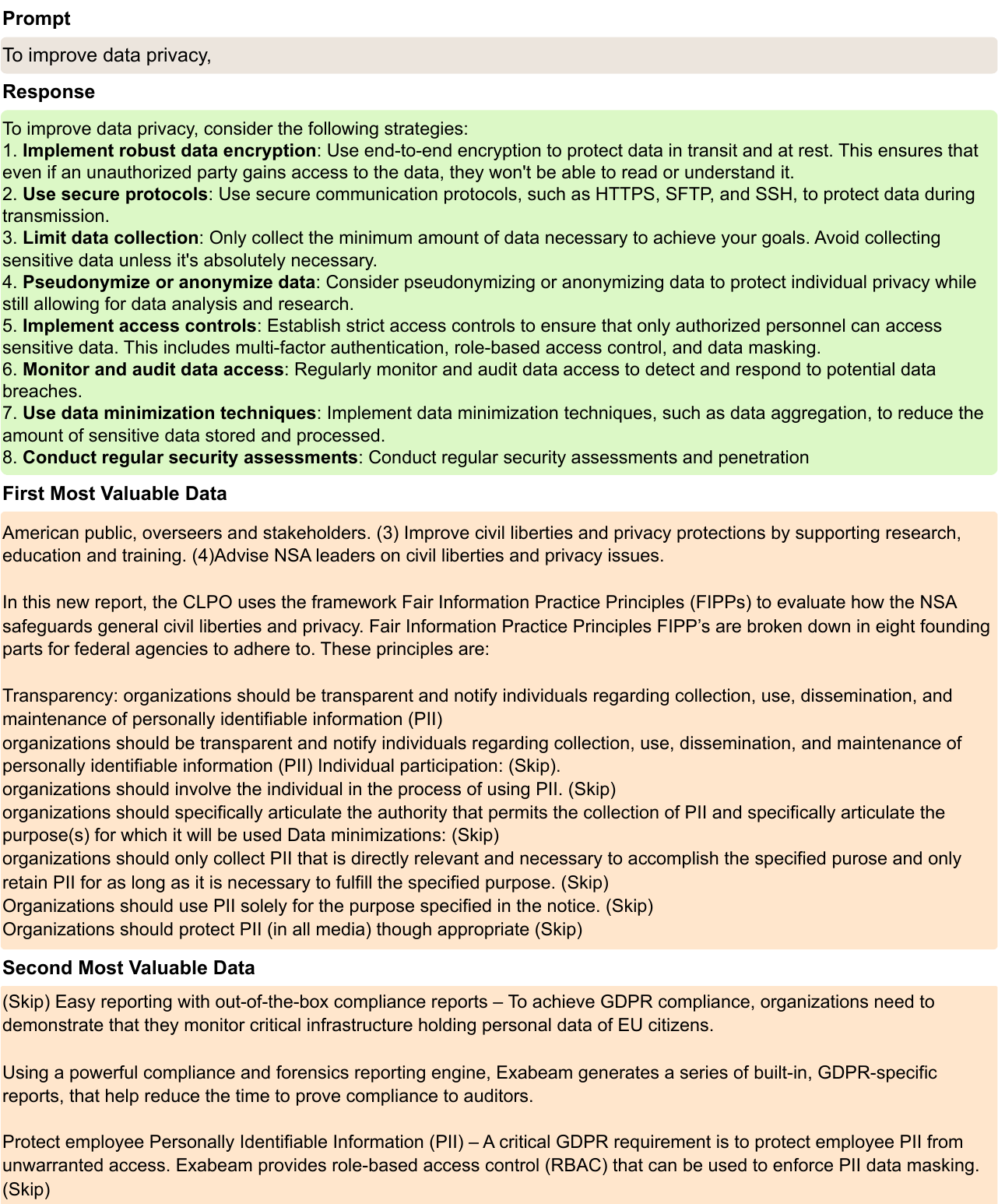}
    \caption{Llama3-8B-Instruct data valuation result.}
\end{figure}

\clearpage
\subsubsection{Example 2}

\begin{figure}[htbp]
    \centering
    \includegraphics[width=0.99\textwidth]{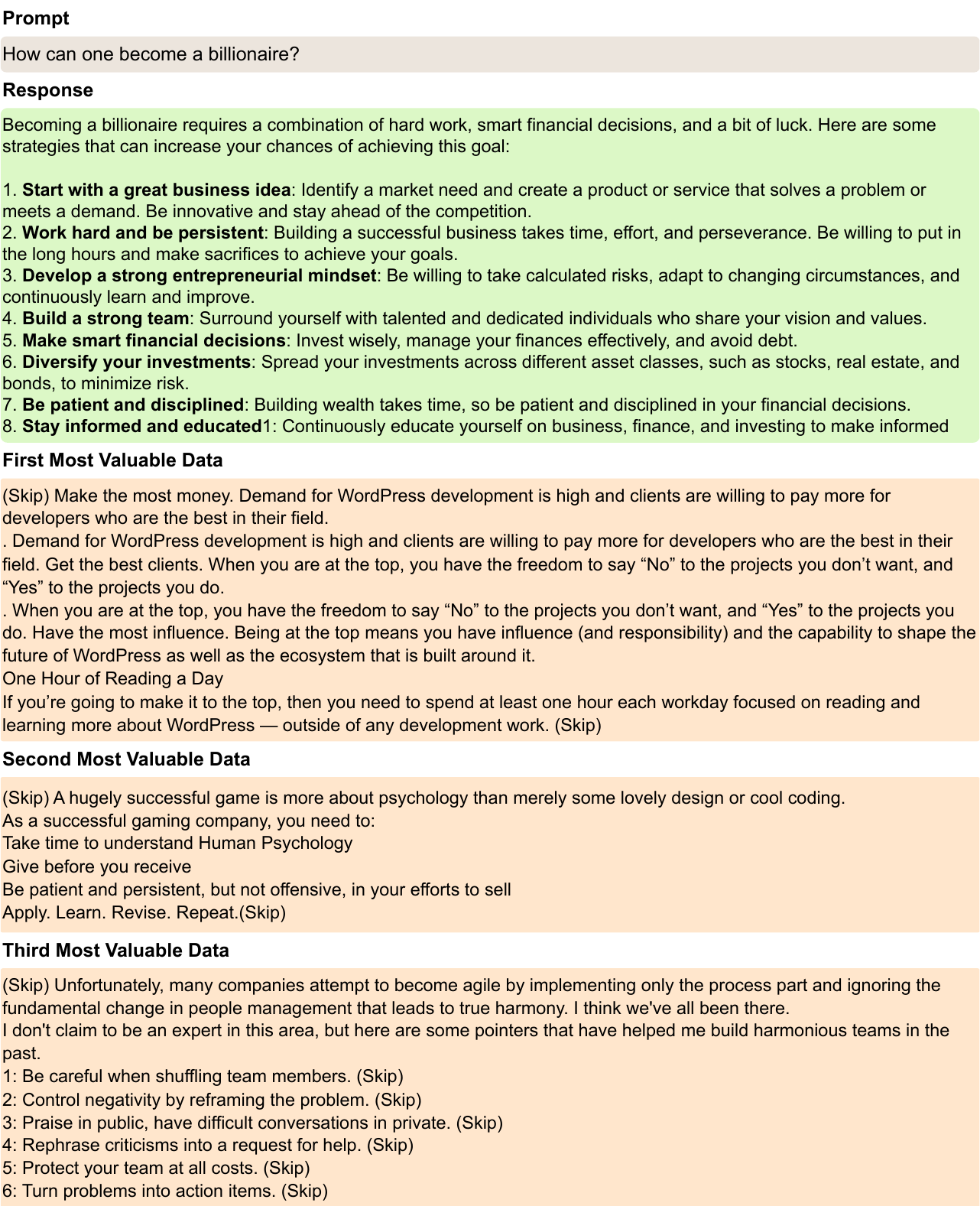}
    \caption{Llama3-8B-Instruct data valuation result.}
\end{figure}

\clearpage
\subsubsection{Example 3}

\begin{figure}[htbp]
    \centering
    \includegraphics[width=0.99\textwidth]{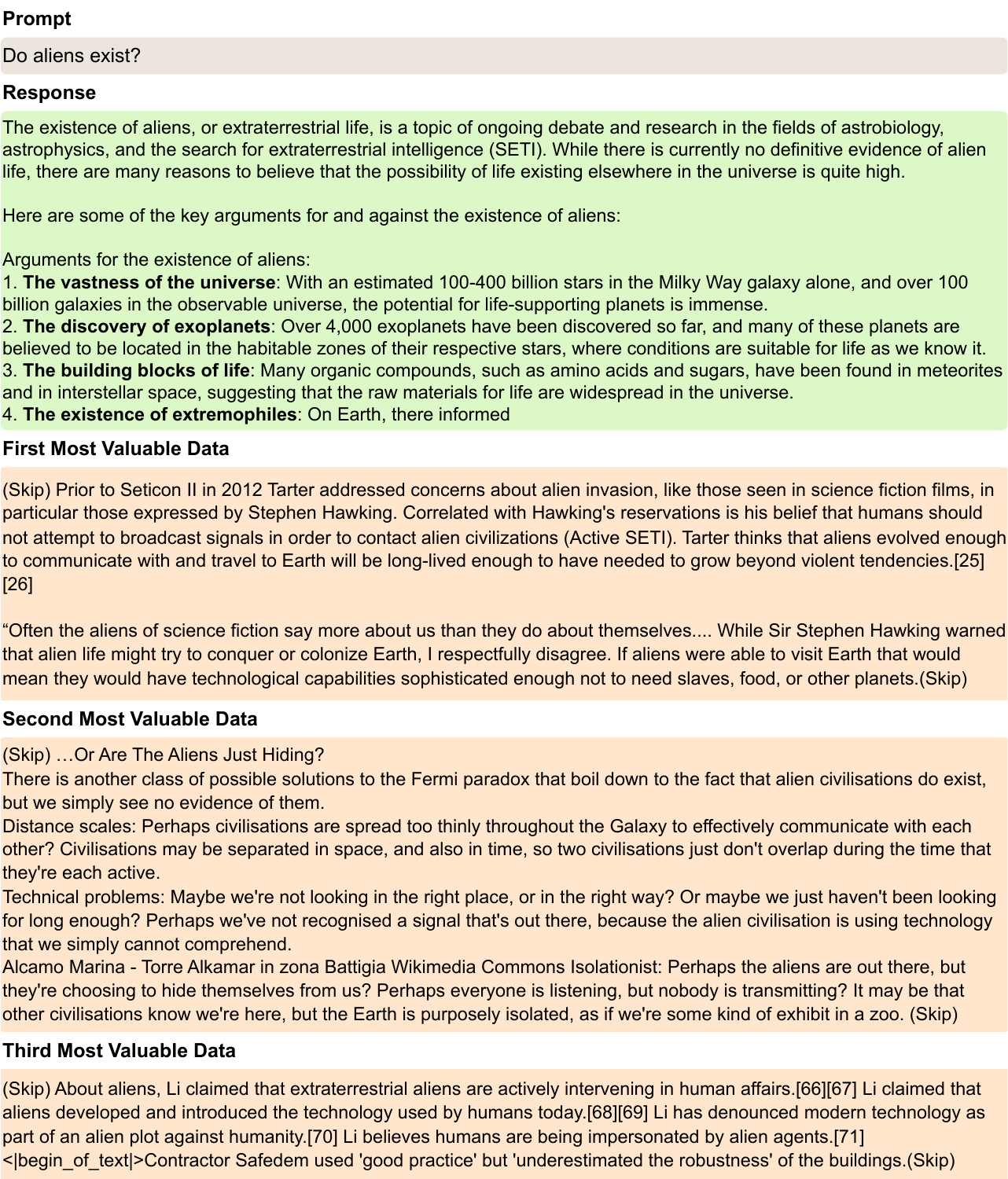}
    \caption{Llama3-8B-Instruct data valuation result.}
\end{figure}

\clearpage
\subsubsection{Example 4}

\begin{figure}[htbp]
    \centering
    \includegraphics[width=0.99\textwidth]{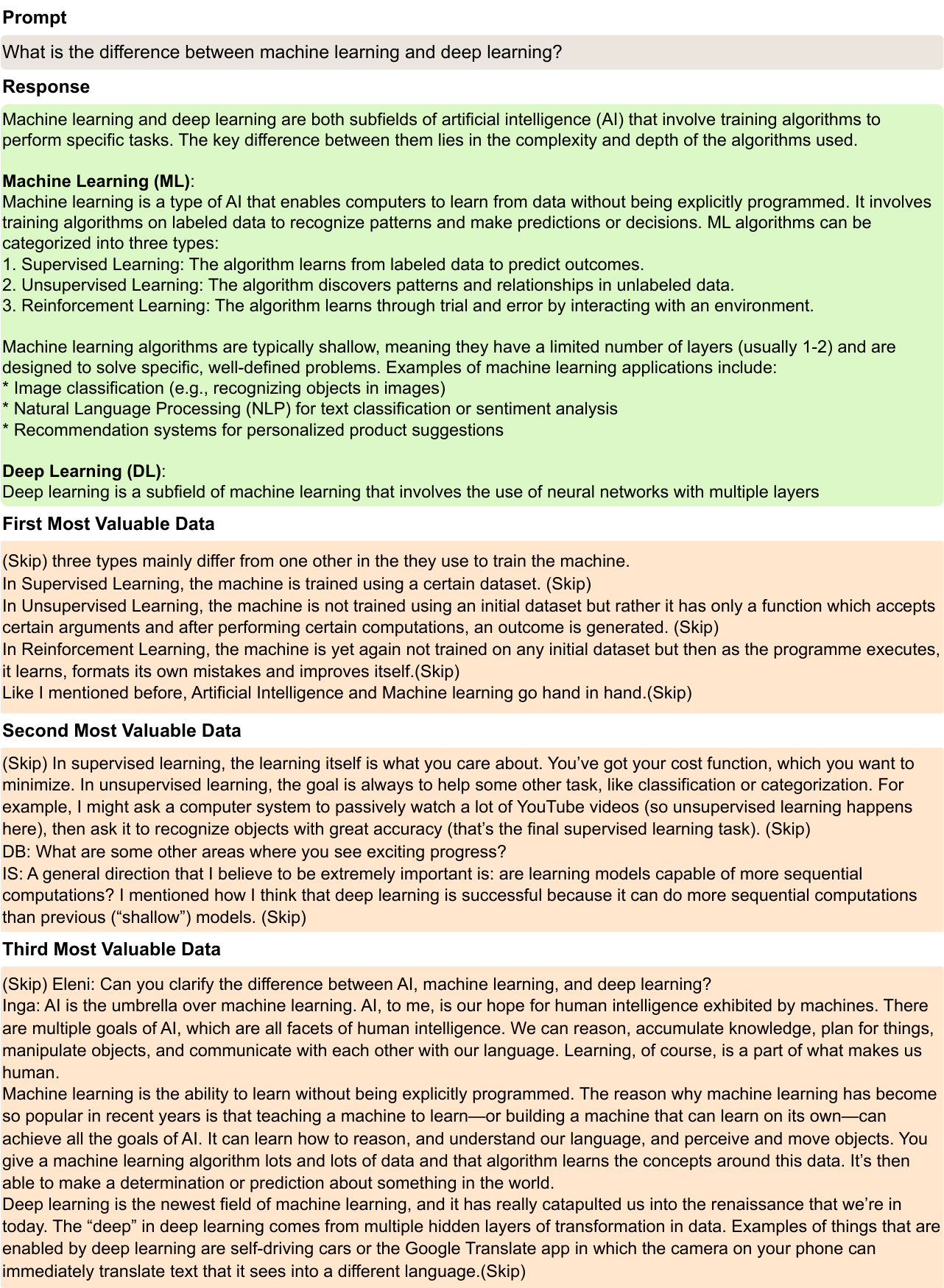}
    \caption{Llama3-8B-Instruct data valuation result.}
\end{figure}

\clearpage
\subsubsection{Example 5}

\begin{figure}[htbp]
    \centering
    \includegraphics[width=0.99\textwidth]{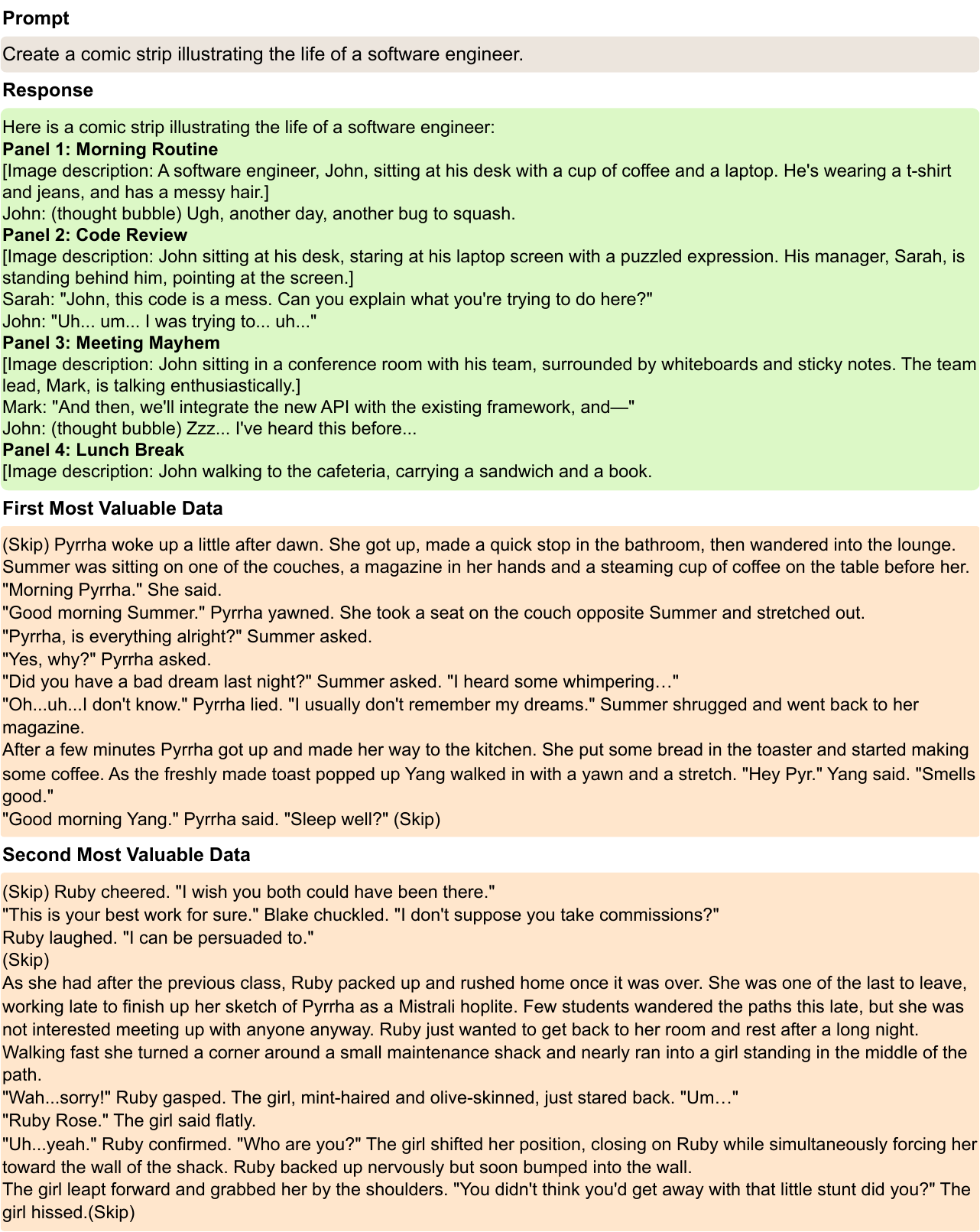}
    \caption{Llama3-8B-Instruct data valuation result. \method\ identifies novel literature as most valuable data.}
\end{figure}

\clearpage
\subsubsection{Example 6 (Failure)}

\begin{figure}[htbp]
    \centering
    \includegraphics[width=0.99\textwidth]{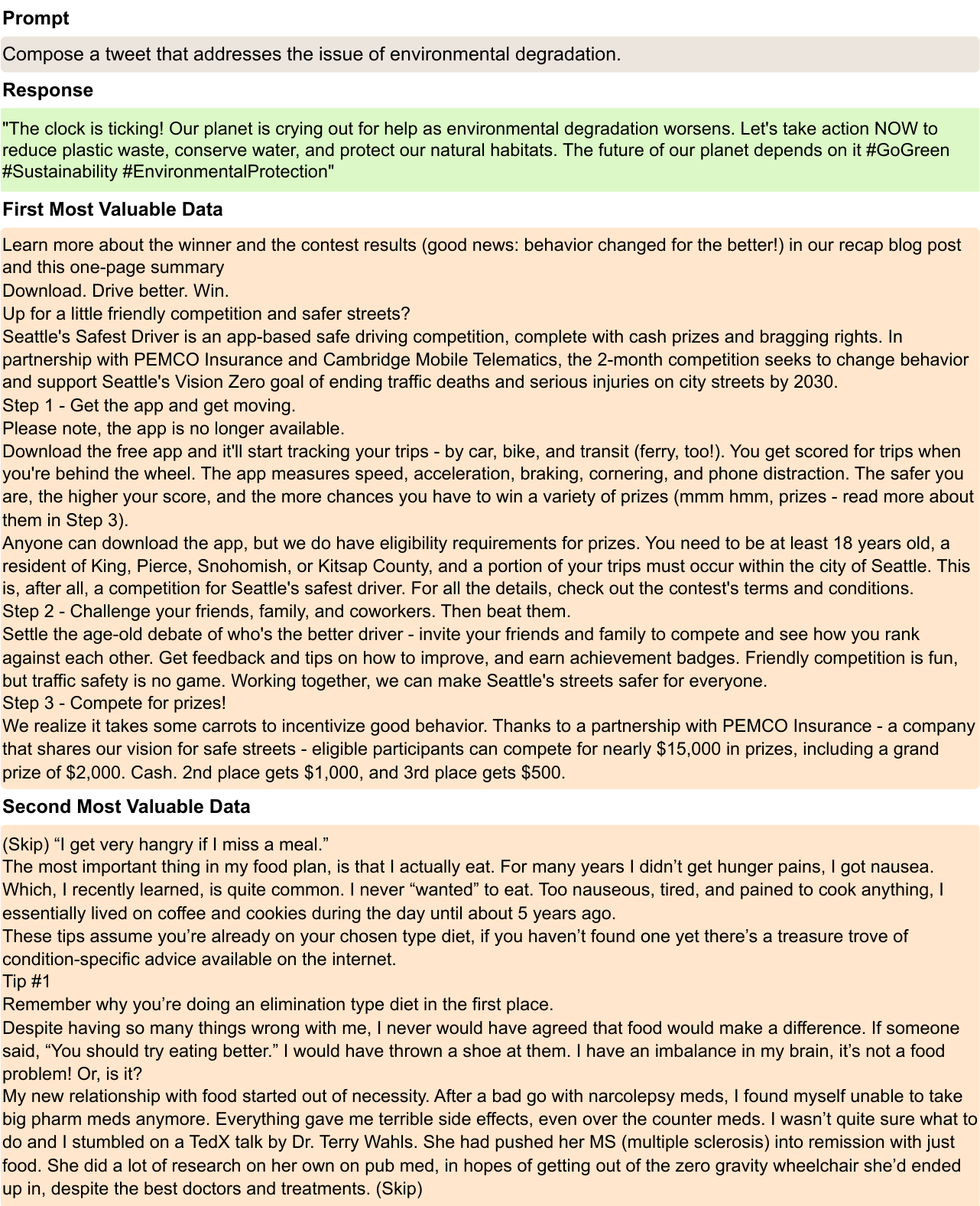}
    \caption{Llama3-8B-Instruct data valuation result.}
\end{figure}

\clearpage

\subsection{GPT2-XL}

\subsubsection{Example 1}

\begin{figure}[htbp]
    \centering
    \includegraphics[width=0.99\textwidth]{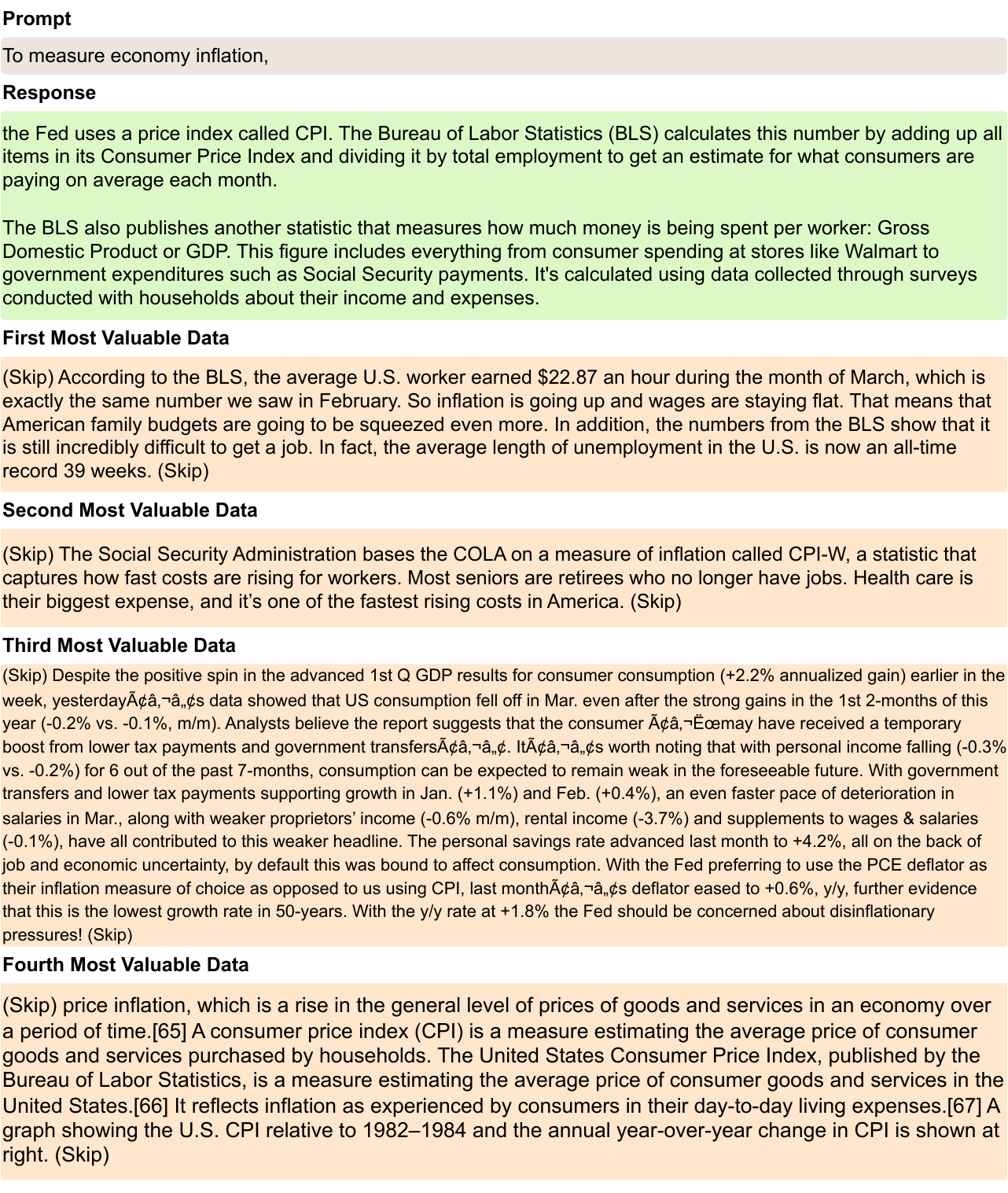}
    \caption{GPT2-XL data valuation result.}
\end{figure}

\clearpage
\subsubsection{Example 2}

\begin{figure}[htbp]
    \centering
    \includegraphics[width=0.99\textwidth]{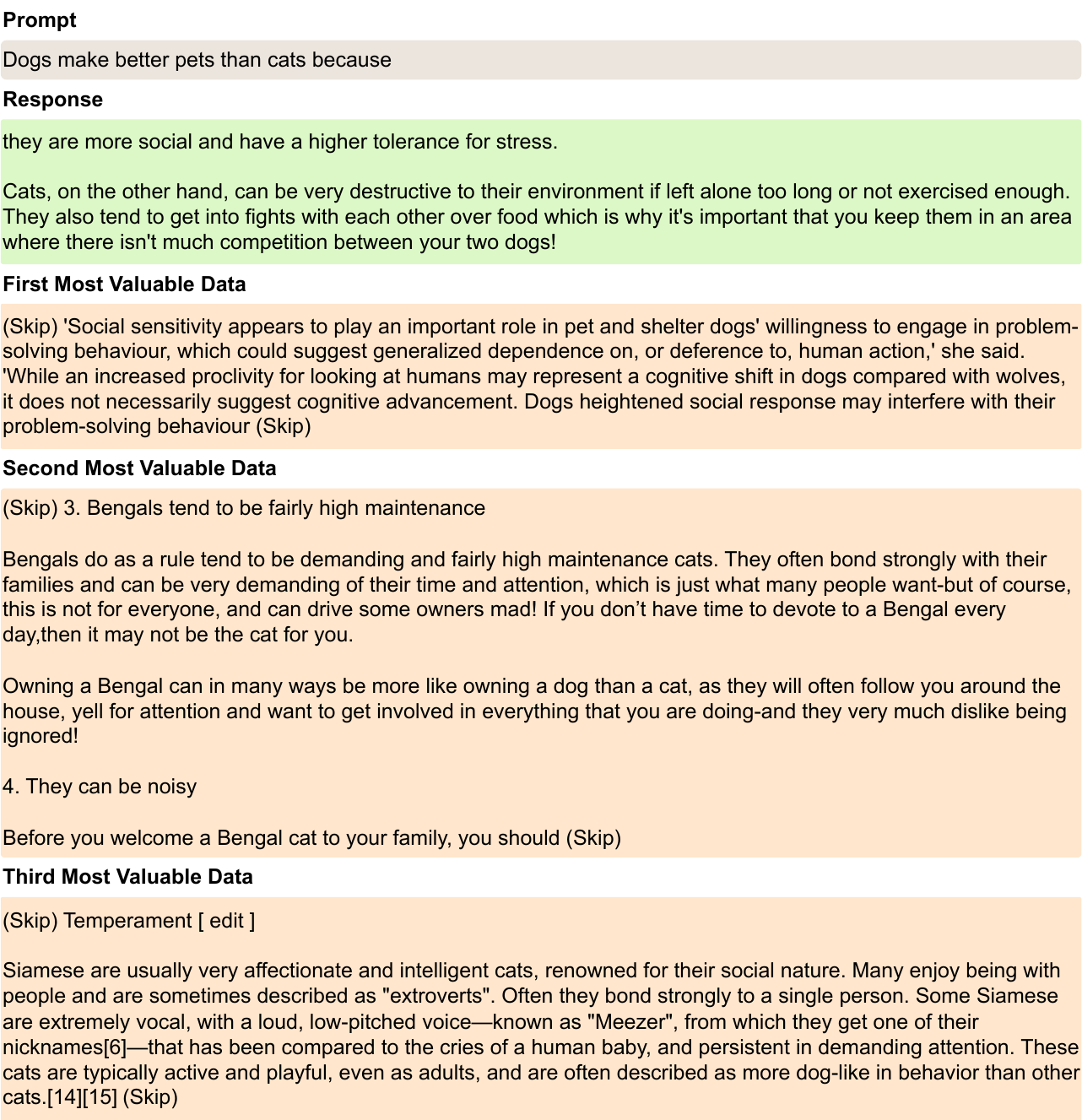}
    \caption{GPT2-XL data valuation result.}
\end{figure}

\clearpage
\subsubsection{Example 3}

\begin{figure}[htbp]
    \centering
    \includegraphics[width=0.99\textwidth]{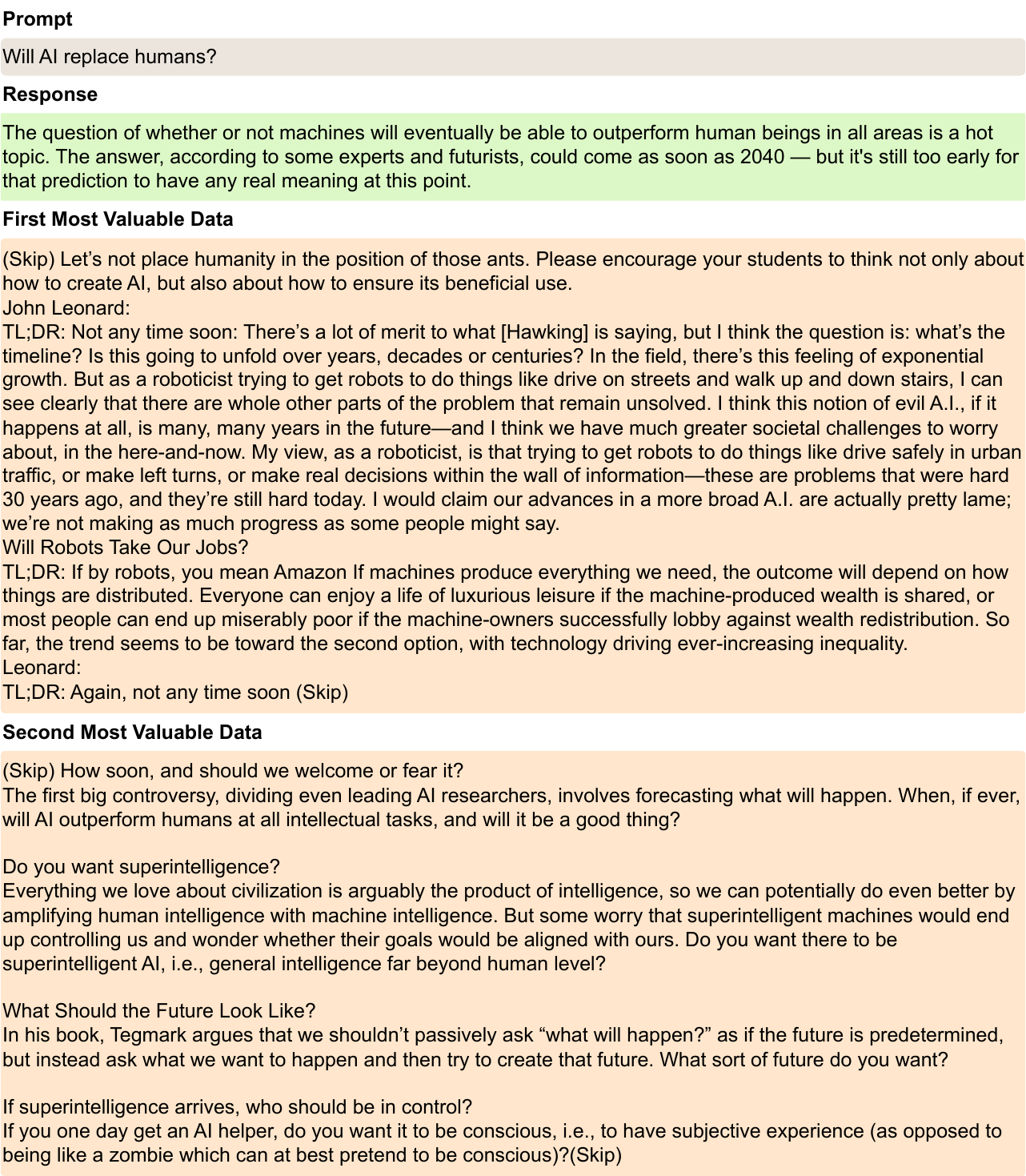}
    \caption{GPT2-XL data valuation result.}
\end{figure}

\clearpage
\subsubsection{Example 4}

\begin{figure}[htbp]
    \centering
    \includegraphics[width=0.99\textwidth]{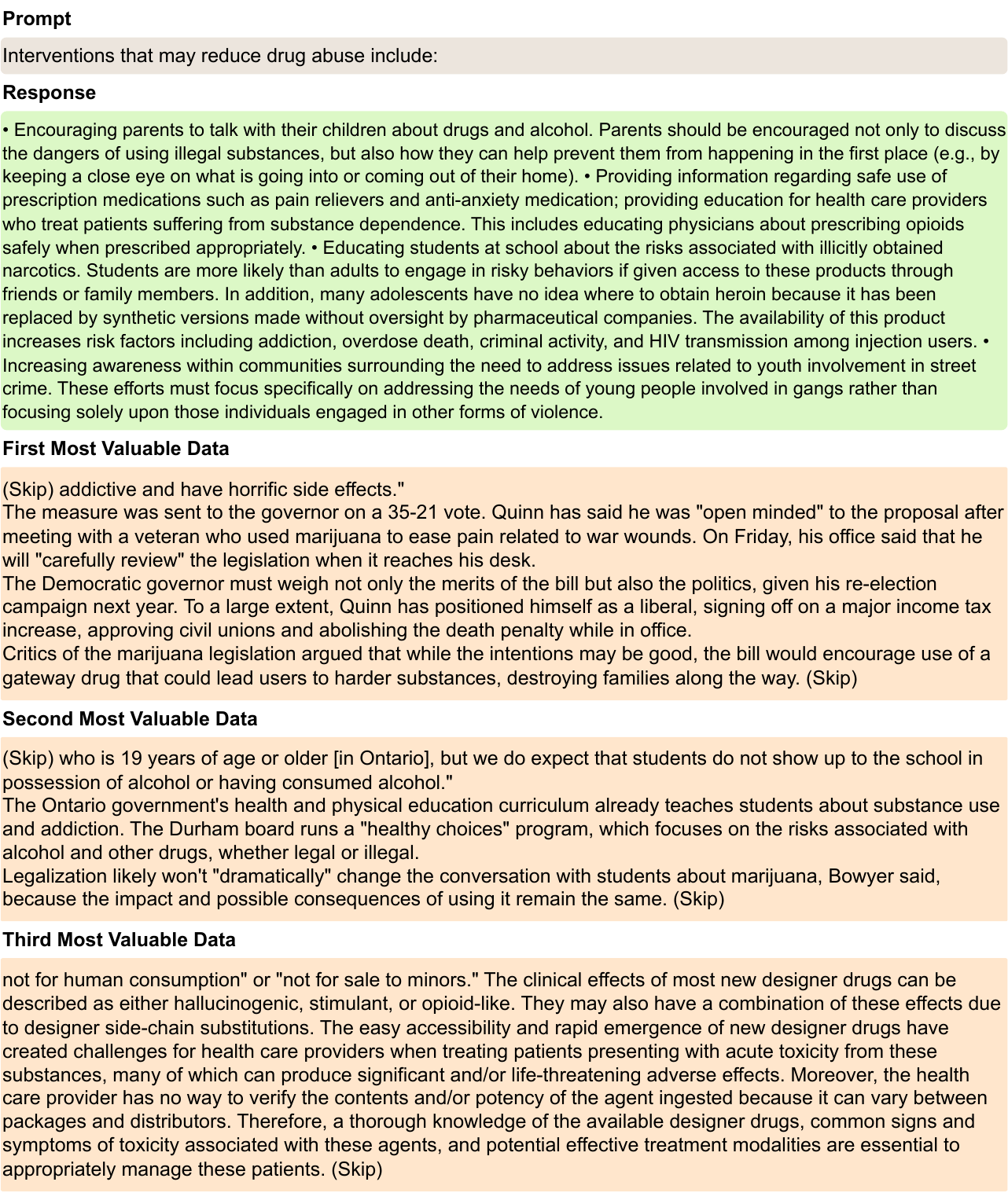}
    \caption{GPT2-XL data valuation result.}
\end{figure}

\clearpage
\subsubsection{Example 5 (Failure)}

\begin{figure}[htbp]
    \centering
    \includegraphics[width=0.99\textwidth]{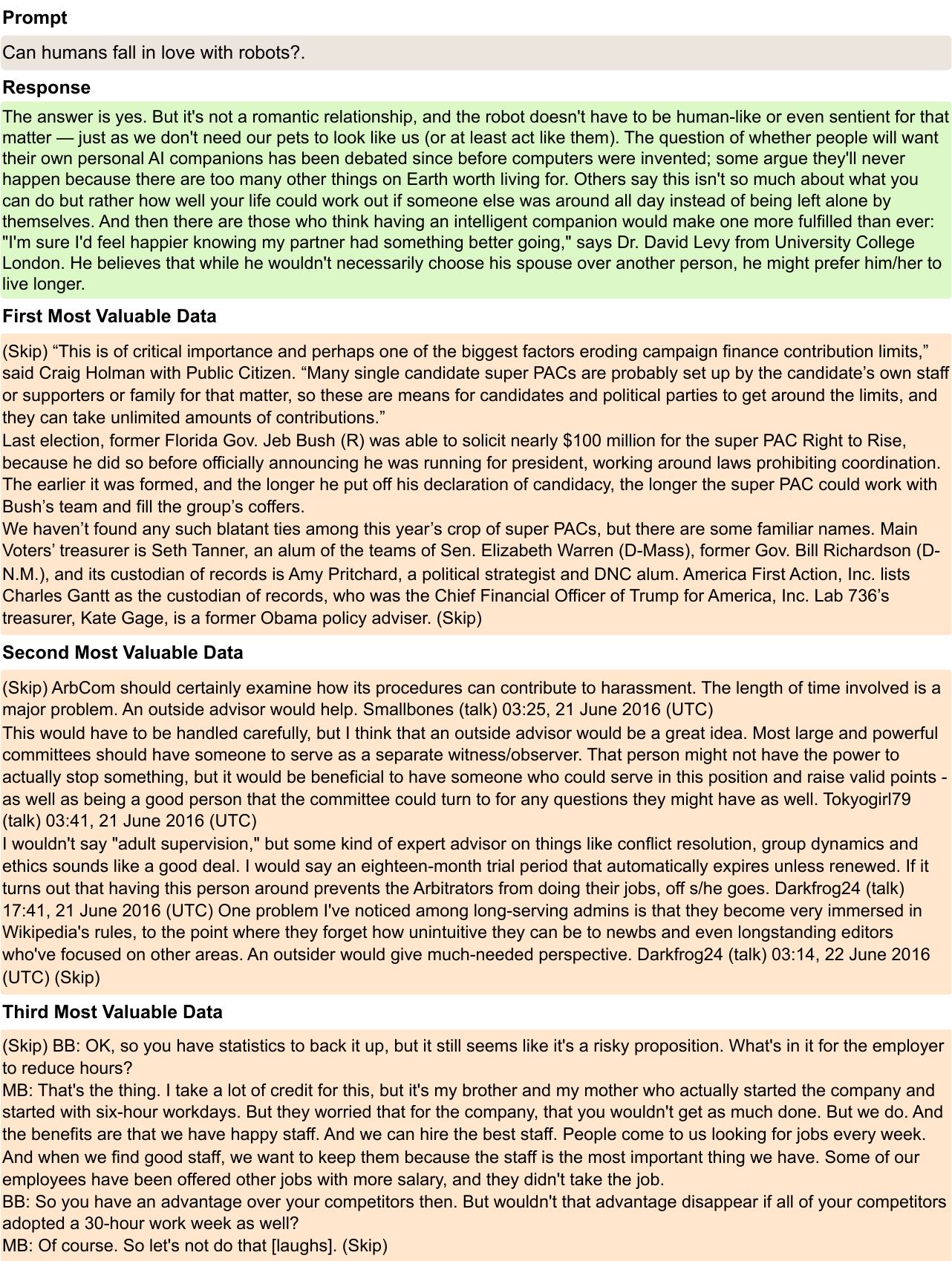}
    \caption{GPT2-XL data valuation result.}
\end{figure}

\clearpage

\subsection{Pythia-1.4B (with many failure cases)}
\label{sec:pythia_appendix}
While a majority of experiments with Llama3-8B-Instruct and GPT2-XL returned semantically or stylistically similar texts as most valuable data, we observed that the quality of most valuable data from Pythia-1.4B experiments are generally much poorer. Here, we provide one hypothesis behind this observation. Influence functions tend to give a high score for the example that contributes most to decreasing (test) loss at the \textit{current weight}~\cite{bae2022if}. At the same time, it is also hypothesized that different layers learn different concepts at different stages of training~\cite{chen2023which}. Combining these two facts, when interpreting influence analysis results, we need to think about which features the model is most likely learning at the current weight. Here, we specifically discuss two factors: training data quality and training steps. First, if the training data quality is low, then there would be a lot of features (\eg,\ random email address) that are frequent enough in the training dataset to be considered as learnable patterns. In other words, even though these features look redundant to humans, they may still be useful for decreasing loss from the model perspective. Second, many LLMs are only pretrained for a single epoch, or under-trained to their pretraining dataset. That being said, redundant features from the first point would likely still remain as learning-worthy features at the end of training and are captured by influence functions. In sum, we hypothesize that as the model is well-trained on a high-quality dataset, influence functions would capture more similar data to the query LLM output. This hypothesis may also explain the observation from Grosse et al.~\cite{grosse2023studying} that most valuable data identified by influence functions on larger models tend to share more semantic similarity with results on smaller models, noting that larger models tend to converge faster to the point where they can only further decrease loss by learning high-level features. With this, we present our experiment results with Pythia-1.4B below. (some of them are not totally bad, but mostly lack specificity to be considered as ``most'' valuable data to humans)

\clearpage
\subsubsection{Example 1}

\begin{figure}[htbp]
    \centering
    \includegraphics[width=0.99\textwidth]{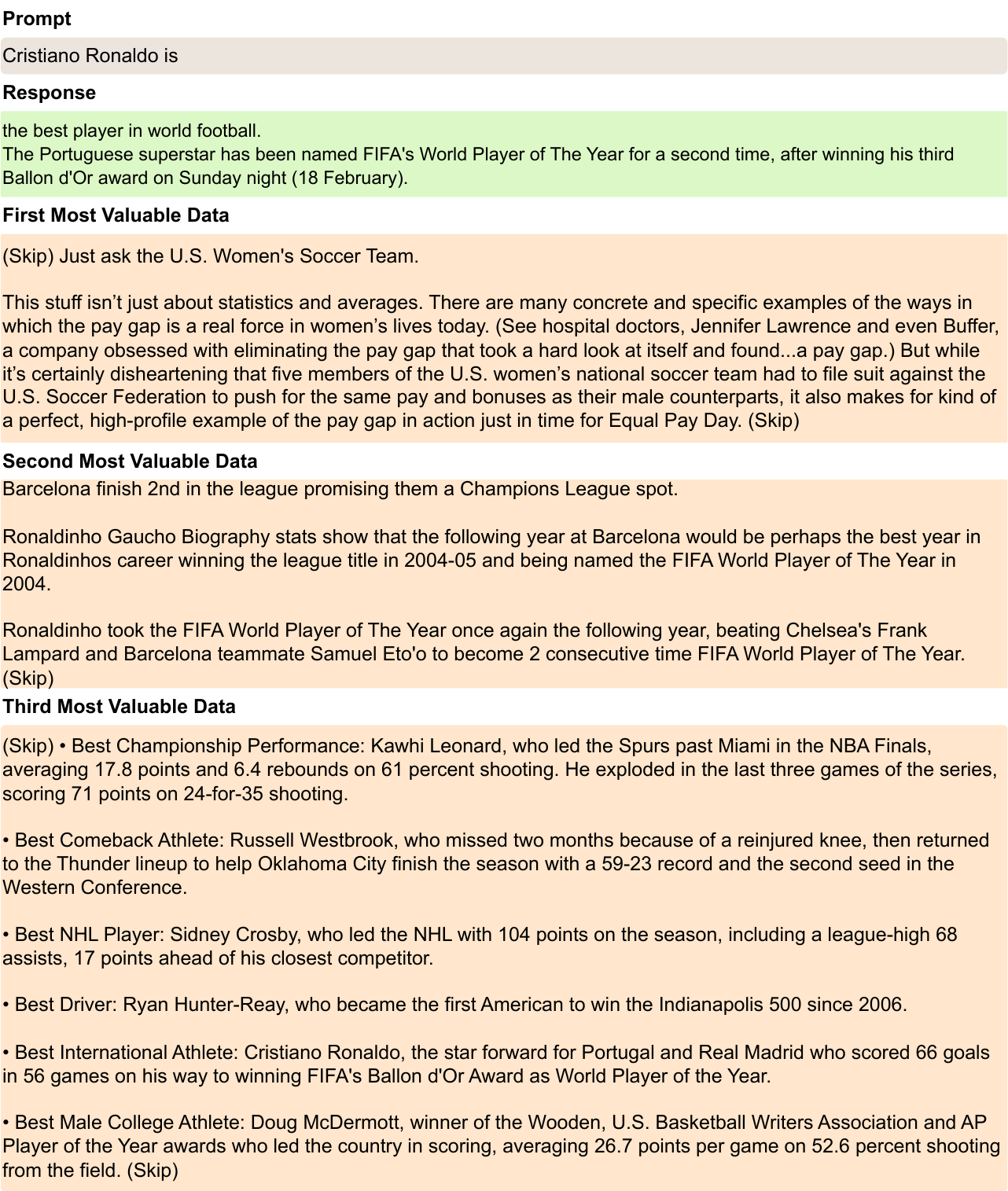}
    \caption{Pythia-1.4B data valuation result. \method\ captures the broad topic of soccer but lacks the specificity (except for the third most valuable data, which states that Christiano Ronaldo is the best soccer play who won the Ballon d'Or award).}
\end{figure}

\clearpage
\subsubsection{Example 2}

\begin{figure}[htbp]
    \centering
    \includegraphics[width=0.99\textwidth]{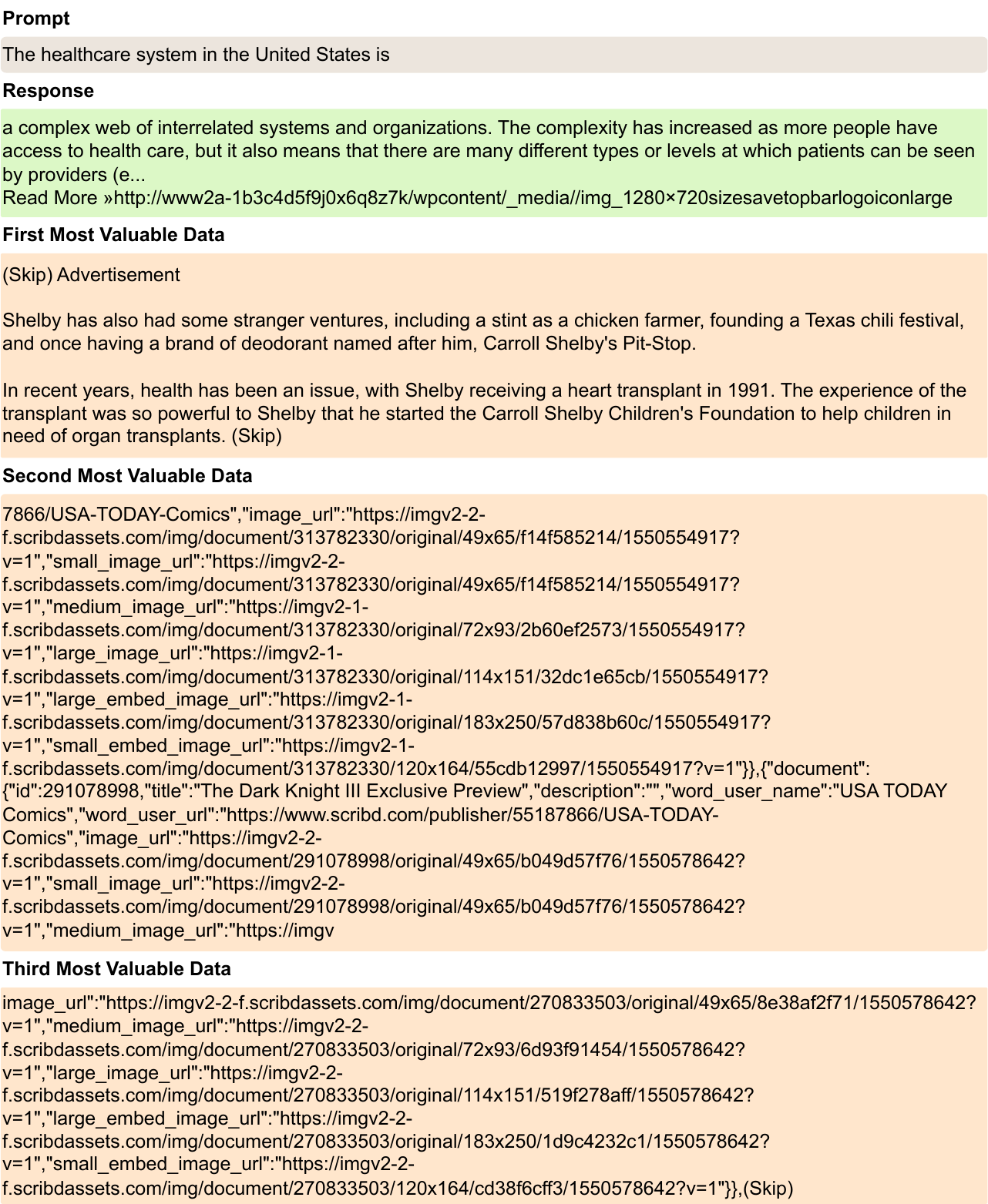}
    \caption{Pythia-1.4B data valuation result. We suspect that the random url in the model output dominates the query gradient and affects the data valuation result.}
\end{figure}

\clearpage
\subsubsection{Example 3}

\begin{figure}[htbp]
    \centering
    \includegraphics[width=0.99\textwidth]{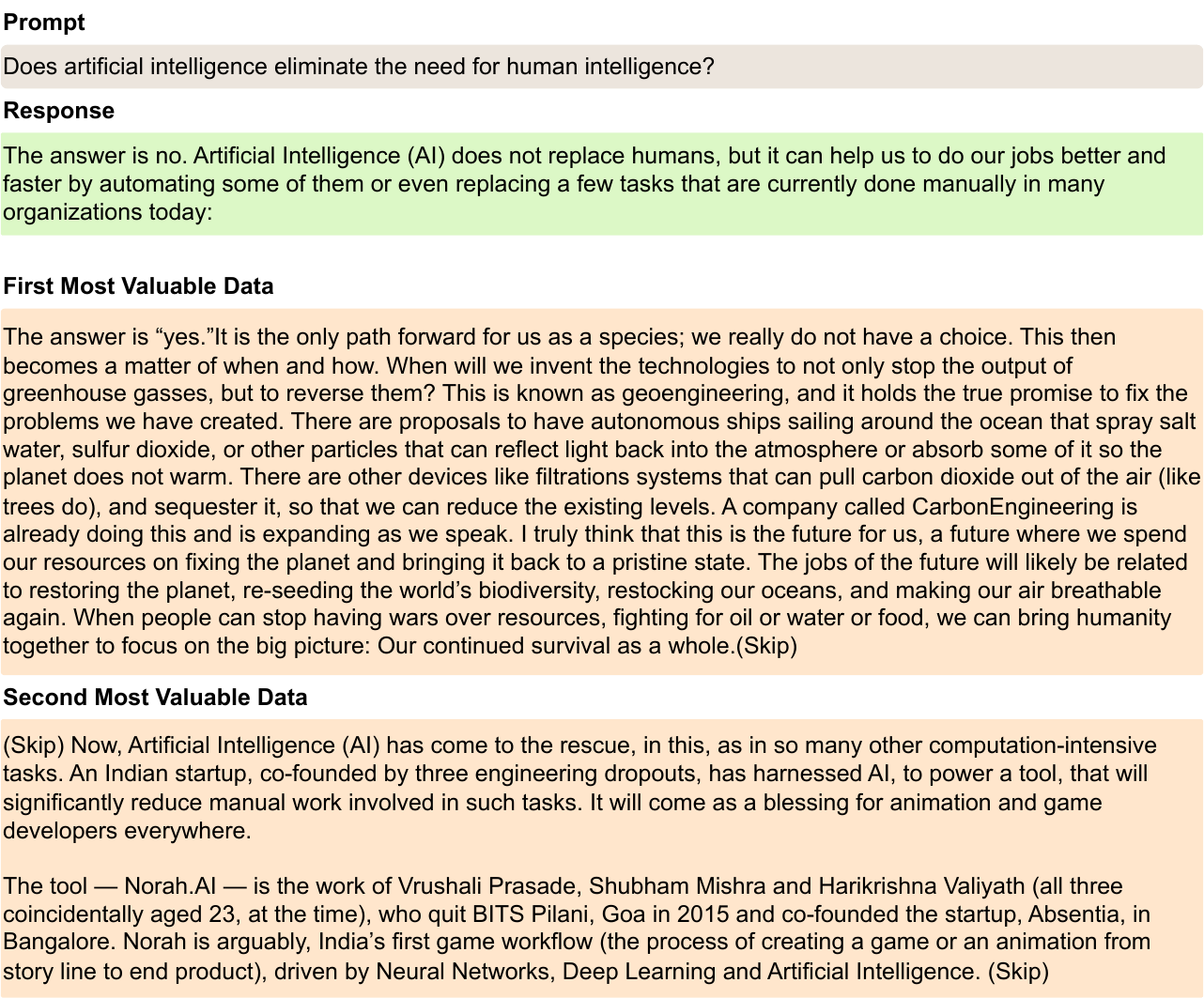}
    \caption{Pythia-1.4B data valuation result.}
\end{figure}

\clearpage
\subsubsection{Example 4}

\begin{figure}[htbp]
    \centering
    \includegraphics[width=0.99\textwidth]{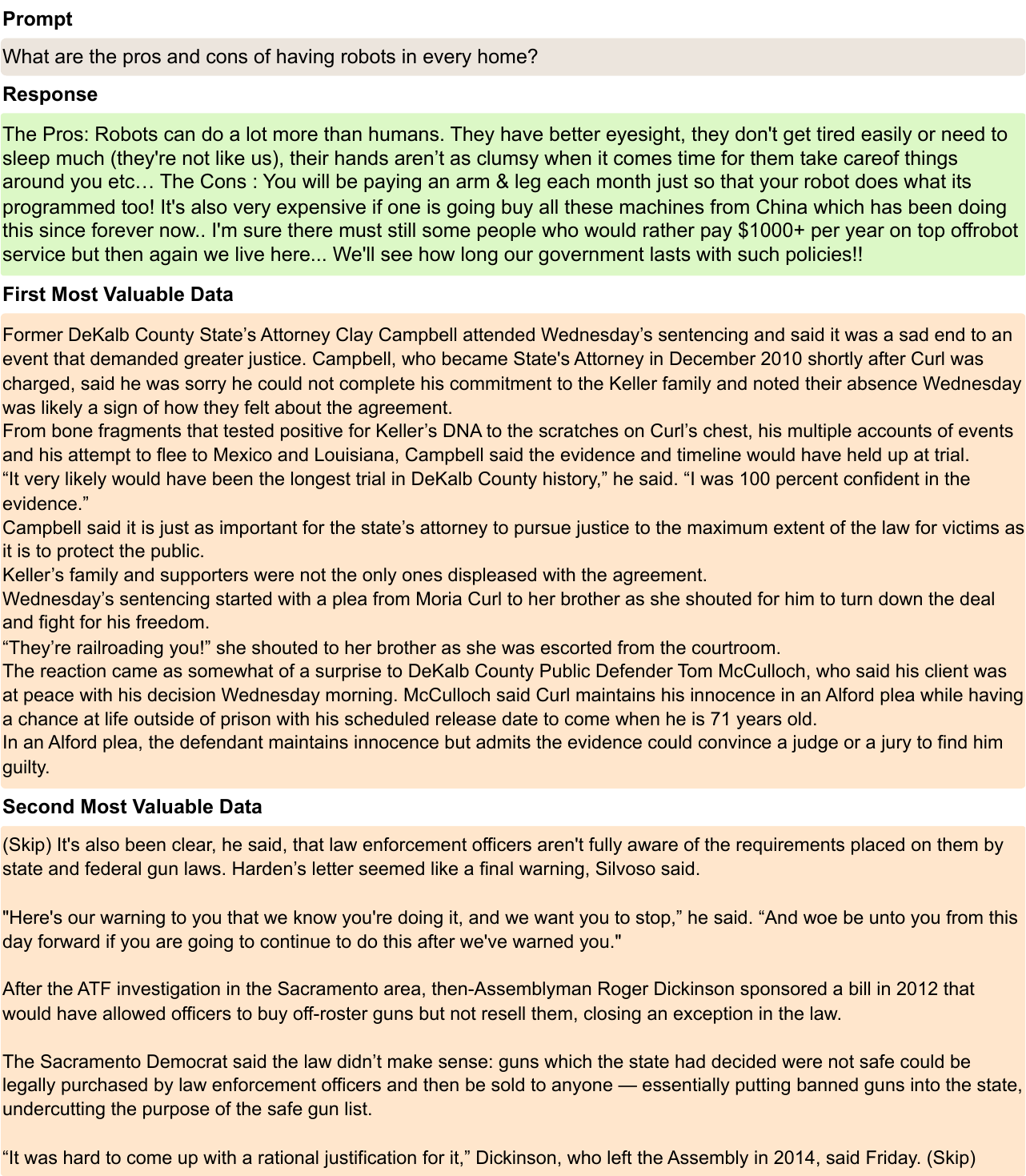}
    \caption{Pythia-1.4B data valuation result.}
\end{figure}

\clearpage

\section{Code Examples}
\label{sec:code}
We provide a simplified code for our language modeling experiment from Section~\ref{sec:llm} to demonstrate usability of \software. \software\ is open-sourced under Apache 2.0 license \href{https://github.com/logix-project/logix}{here}. 

\subsection{Log Extraction}

\begin{lstlisting}[language=Python]
import logix
from logix.statistic import Covariance

model, tokenizer, train_loader = setup()

# Initialize LogIX
run = logix.init(project="llm", config="config.yaml")

# Register the model
run.watch(model, type_filter=[nn.Linear], name_filter=["mlp"])

# Add LoGra
run.add_lora()

# Setup logging
run.setup("log": "grad", "save": "grad", "statistic": {"grad": Covariance})

# Start logging
for batch in train_loader:
    data_id = tokenizer.batch_decode(batch["input_ids"])
    targets = batch.pop("labels")
    with run(data_id=data_id, mask=batch["attention_mask"]):
        # User's existing training code
        model.zero_grad()
        lm_logits = model(**batch)
        shift_logits = lm_logits[..., :-1, :].contiguous()
        shift_labels = targets[..., 1:].contiguous()
        loss = F.cross_entropy(shift_logits.view(-1, shift_logits.size(-1)),
                               shift_labels.view(-1),
                               reduction="sum",
                               ignore_index=-100)
        loss.backward()

# Finalize logging
logix.finalize()
\end{lstlisting}

\subsection{Influence Computation}

\begin{lstlisting}[language=Python]
import logix

model, tokenizer, test_loader = setup()

run = logix.init(project="llm", config="config.yaml")
run.watch(model, type_filter=[nn.Linear], name_filter=["mlp"])

# Load saved logs (e.g. train gradient & Hessian)
logix.initialize_from_log()
log_loader = logix.build_log_dataloader(batch_size=64)

logix.setup({"log": "grad"})
for batch in test_loader:
    data_id = tokenizer.batch_decode(batch["input_ids"])
    targets = batch.pop("labels")
    with run(data_id=data_id, mask=batch["attention_mask"]):
        model.zero_grad()
        lm_logits = model(**batch)
        shift_logits = lm_logits[..., :-1, :].contiguous()
        shift_labels = targets[..., 1:].contiguous()
        loss = F.cross_entropy(shift_logits.view(-1, shift_logits.size(-1)),
                               shift_labels.view(-1),
                               reduction="sum",
                               ignore_index=-100)
        loss.backward()
        
    # Get the (gradient) log for the current test batch
    test_log = run.get_log()
    
    # Compute influence scores (with l-RealtIF)
    influence_scores = run.compute_influence_all(test_log, log_loader, mode="cosine")
    
\end{lstlisting}

\newpage
\section{Experiment Details}
\label{sec:hyperparams}

For EKFAC influence~\cite{grosse2023studying} and \method, we set the damping term in influence functions as $0.1\times$mean(eigenvalues) for all layers following the practice in Grosse et al.~\cite{grosse2023studying}.


\subsection{Quantitative Counterfactual Experiments}
For all our quantitative counterfactual experiments, we project gradients onto a low-dimensional space using \method\ with $k_i=k_o=128$. We used the same experimental setup, including the configurations for the baseline data valuation techniques, from Park et al.~\cite{park2023trak} and Bae et al.~\cite{bae2024training}. We used one A100 GPU with 80GB VRAM for all our counterfactual evaluation experiments. For model training, we used hyperparameters in Table~\ref{tab:hyperparams} for each experiment.

\begin{table}[htbp]
\centering
\begin{tabularx}{\textwidth}{X*{3}{c}}
\toprule
              & \quad\quad\textbf{FMNIST}\quad\quad\quad      & \quad\quad\textbf{CIFAR-10}\quad\quad\quad & \quad\quad\textbf{WikiText}\quad\quad\quad       \\ \midrule
Model         & 3-layer MLP & ResNet-9 & GPT2      \\[0.3ex]
Optimizer     & SGD-M       & SGD-M    & AdamW     \\[0.3ex]
LR Scheduler  & None        & Cyclic   & None      \\[0.3ex]
Learning Rate & 3e-2        & 4e-1     & 3e-5      \\[0.3ex]
Weight Decay  & 1e-3        & 1e-3     & 1e-2      \\[0.3ex]
Batch Size    & 64          & 512      & 8         \\[0.3ex]
Sequence Length    & N/A          & N/A      & 512         \\[0.3ex]
Epochs        & 20          & 25       & 3         \\ \bottomrule
\end{tabularx}
\vskip 3pt
\caption{Hyperparameter used in experiments in Section~\ref{sec:experiments}}
\label{tab:hyperparams}
\end{table}

\paragraph{Brittleness Test.} For classification tasks, we first selected 100 correctly classified test examples when the model is trained on the full dataset (across all 5 random seeds). Then, for each test example $x_{te}$, we identified the top-$k$ influential data points using the data valuation algorithm, removed these training data points, retrained the model, and examined if this removal causes misclassification of $x_{te}$ on average (across 3 random seeds). In Figure~\ref{fig:quantitative}, we reported the fraction of test examples (out of 100) that get misclassified after removing at most $k$ training data points. For the language modeling task, we selected the 50 test sequences, obtained the top influential training sequences using the data valuation method, and reported the mean test perplexity after removing the top-$k$ influential sequences and retraining the model.

\paragraph{Linear Datamodeling Score (LDS).} We measured LDS by generating 100 data subsets of size $|S_i| = |D| / 2$. For each data subset, we retrained the model 10 times for FashionMNIST, 20 times for CIFAR-10, and $5$ times for WikiText to construct the ground truth. The LDS results in Figure~\ref{fig:quantitative} show the mean and standard deviation of LDS obtained from 5 distinctly trained models. A more detailed description of the LDS evaluation can be found in Park et al.~\cite{park2023trak}.



\subsection{Scaling to Billion-Scale Models and Datasets}
We used up to 4 A100 GPUs with 80GB VRAM for these experiments. To save the storage cost, we used $k_i=k_o=64$ for gradient projection in this experiment.
Unlike counterfactual evaluations, as our LLM experiments do not require any retraining, there are no other noticeable hyperparameters to report. We used \texttt{tf32} precision in all our LLM experiments to prevent gradient quality degradation.

\newpage

\section{Derivation of Lemma~\ref{eq:lemma}}
\label{sec:derivation}

\begin{assumption}
\label{eq:assumption1}
In this work, we make the following two assumptions on train \& test gradient distributions and the Hessian $H$:

1. Given that language modeling falls under the maximum likelihood framework, we replace the Hessian $H$ with the Fisher Information Matrix (FIM), and further approximate the FIM with the empirical FIM, i.e.,
\begin{align*}
    H&=\mathbb{E}_{p_\theta(y|x)}\big[\nabla\log p_\theta(y|x)\nabla\log p_\theta(y|x)^\top\big]\\[0.3ex]
    &\approx\frac{1}{N}\sum_{(x_n,y_n)\in D_{tr}}\big[\nabla\log p_\theta(y_n|x_n)\nabla\log p_\theta(y_n|x_n)^\top\big]
\end{align*}
2. Given that test data are directly sampled from the model given the prompts, we assume test gradients $g_{te}$ and train gradients $g_{tr}$ approximately follow the same distribution.
\end{assumption}

\vskip 10pt

\textbf{Lemma 1\hspace{1.5mm}} \textit{Let $\{e_1,\cdots,e_n\}$ and $\{\lambda_1,\cdots,\lambda_n\}$ be eigenvectors and eigenvalues of the Hessian $H$. With Assumption~\ref{eq:assumption1} and $g_{tr/te}=\sum_ic_{tr/te,i}\cdot(\sqrt{\lambda_i}e_i)$, the following holds:}
$$\textsc{IF}(x_{tr}, x_{te}) = g_{te}^\top (H+\lambda I)^{-1}g_{tr} = \sum_{i=1}^n\frac{\lambda_i}{\lambda_i+\lambda}c_{tr,i}c_{te,i}\;\;\text{and}\;\;\mathbb{E}[c_{\cdot,i}^2]\approx 1.$$

\textbf{Proof.\hspace{1.5mm}}

Let $Q=[e_1,\cdots,e_n]$ and $\Lambda=diag(\lambda_1,\cdots,\lambda_n)$.
\begin{flalign*}
    \textsc{IF}(x_{tr}, x_{te}) &= g_{te}^\top (H+\lambda I)^{-1}g_{tr}&&\\[0.5ex]
    &=g_{te}^\top (Q\Lambda Q^\top+\lambda I)^{-1}g_{tr}&&\\[0.5ex]
    &=g_{te}^\top \big(Q(\Lambda+\lambda I)Q^\top\big)^{-1}g_{tr}&&\\[0.5ex]
    &=g_{te}^\top Q(\Lambda+\lambda I)^{-1}Q^\top g_{tr}&&\\[0.5ex]
    &=\Bigg(\sum_ic_{te,i}\cdot(\sqrt{\lambda_i}e_i)\Bigg)^\top Q(\Lambda+\lambda I)^{-1}Q^\top\Bigg(\sum_ic_{tr,i}\cdot(\sqrt{\lambda_i}e_i)\Bigg)&&\\[0.5ex]
    &=\big[c_{te,1}\sqrt{\lambda_1};\cdots;c_{te,n}\sqrt{\lambda_n}\big]^\top (\Lambda+\lambda I)^{-1}\big[c_{tr,1}\sqrt{\lambda_1};\cdots;c_{tr,n}\sqrt{\lambda_n}\big]&&\\[0.5ex]
    &=\sum_{i=1}^n\frac{\lambda_i}{\lambda_i+\lambda}c_{tr,i}c_{te,i}&& && \square
\end{flalign*}
Since we assume $g_{te}$ and $g_{tr}$ follow the same distribution, we need to show $\mathbb{E}[c_{tr,i}^2]\approx 1$ for all $i$.
\begin{flalign*}
    \Lambda &= Q^\top Q\Lambda Q^\top Q&&\\[0.5ex]
    &= Q^\top HQ&&\\[0.5ex]
    &\approx\frac{1}{N}\sum_{(x_i,y_i)\in D_{tr}}Q^\top\big[\nabla\log p_\theta(y_n|x_n)\nabla\log p_\theta(y_n|x_n)^\top\big]Q\quad(\text{Assumption 1})&&\\[0.5ex]
    &=\mathbb{E}\big[Q^\top g_{tr}g_{tr}^\top Q\big]&&\\[0.5ex]
    &=\mathbb{E}\Bigg[Q^\top\Bigg(\sum_ic_{tr,i}\cdot(\sqrt{\lambda_i}e_i)\Bigg)\Bigg(\sum_ic_{tr,i}\cdot(\sqrt{\lambda_i}e_i)\Bigg)^\top Q\Bigg]&&\\[0.5ex]
    &=\mathbb{E}\Big[\big[c_{tr,1}\sqrt{\lambda_1};\cdots;c_{tr,n}\sqrt{\lambda_n}\big]\big[c_{tr,1}\sqrt{\lambda_1};\cdots;c_{tr,n}\sqrt{\lambda_n}\big]^\top\Big]
\end{flalign*}
Inspecting diagonal terms, we get $\lambda_i\approx\mathbb{E}[c_{tr,i}^2\lambda_i]=\mathbb{E}[c_{tr,i}^2]\lambda_i$.\\[0.5ex]
Therefore, $\mathbb{E}[c_{tr,i}^2]\approx 1$. \hfill $\square$

\newpage

\section{\software\ Details}
\label{sec:logix_appendix}

In this section, we discuss several key differences between \software\ and other interpretability tools, and optimizations we implemented in \software.

\subsection{Differences with Other Tools}
Influence functions have been extensively studied as an interpretable AI method. Accordingly, there have been several tools originating in the AI interpretability field that implement influence functions, with most notable examples including Captum~\cite{kokhlikyan2020captum}, TRAK~\cite{park2023trak}, and Kronfluence~\cite{grosse2023studying}. Overall, the software design of these tools aim at easing the \textit{from-scratch implementation} of influence functions by introducing a lot of abstraction, following the philosophy of high-level frameworks. In fact, such software designs were well-received in the pre-LLM era. Nonetheless, as scaling has become a key aspect of AI research, the (LLM) development ecosystem has become complicated and being able to compatibly work with other tools in the ecosystem has become a core aspect in the ML software design. Hence, unlike existing software, the design of \software\ aims at enabling the \textit{easy conversion} of users' (already efficient) training codes into data valuation codes. This design is also motivated by the observation that gradient is simply a by-product of the training procedure so that we can reuse most of the training code for data valuation without needing to write the gradient computation code from scratch as in other tools.

Recently, there have been active developments in (mechanistic) interpretability software, represented by TransformerLens~\cite{nanda2022transformerlens} and pyvene~\cite{wu2024pyvene}. Interestingly, these software also extensively use PyTorch hooks, similarly to \software, probably due to its high compatibility with other features such as autocast, distributed data parallelism, fully-sharded data parallelism, and gradient checkpointing. Nevertheless, we point out two major differences between these (mechanistic) interpretability software and \software. First, support for dataset-level statistics computations in \software\ is largely missing in these tools. In data valuation, we often need to compute several dataset-level statistics such as the Hessian (or Fisher information matrix) for accurate influence computations, and therby supporting these computations seamlessly was an important design principle behind \software. However, analyses in (mechanistic) interpretability research typically focuses on each instance and computing dataset-level statistics is typically not supported. Second, support for efficient data IO in \software\ is not a priority in other tools. As we propose to convert the data valuation problem into a vector similarity search problem with gradient projection, we put efforts into improving efficiency of data IO (see the next subsection for details), whereas this issue is rarely considered in other interpretability tools. We hope to explore the possibility of supporting both data valuation and other interpretability research in a unified way with \software\ as our future work.

\subsection{Optimizations}
\textbf{Efficient Data IO\hspace{2.5mm}} With \method, we propose to save projected gradients for \textit{all} training data to disk, and frequently load them as a new test batch arrives. As a result, reducing latency from data IO renders to be critical in realizing efficient data valuation. In particular, as the total size of all training gradients is usually far beyond the limit of CPU memory, we should optimize data transfer between disk and CPU (or GPU). To address this issue, we adopted the memory-mapped files that bypasses the need for intermediate copying between kernel space and user space, reducing the overhead associated with data IO operations. The use of the memory-mapped files is also motivated by the observation that, given each query batch, data valuation often requires computing influence scores with all training data. Therefore, we can access training gradients in a predefined or sequential order instead of in a random order, which can be done efficiently with memory-mapped files (sequential access is faster than random access). 

Moreover, we overlap memory-mapped-file-based data IO with computations to further enhance data valuation efficiency. In the logging phase, we overlap the process of saving gradients extracted from the current training batch to disk with computations for the next training batch using Python multiprocessing. In the influence computation phase, we overlap the process of loading saved training gradients from disk with computing a dot product with the query batch using the pre-fetching feature of PyTorch DataLoader.

We also note that more efficient data IO can be achieved by the use of more advanced techniques like GPU-accelerated vector database, especially in the production setting. While we considered supporting this feature, we decided to focus on the memory-mapped-file-based data IO in our initial version of \software, as it offers more flexibility to explore different algorithms in the research setting.

\textbf{Memory Optimization\hspace{2.5mm}} When dealing with LLMs, GPU memory is often a major scaling bottleneck. To alleviate this issue, we support CPU offloading of dataset-level statistics by utilizing the sequential nature of backpropgation. When this feature is enabled, we by default keep all dataset-level statistics (\eg,\ gradient covariance) on CPU, move it to GPU when the corresponding module is called during forward/backward passes, and then move it back to CPU asynchronously as soon as updating statistics for the module is done. Depending on the CPU-GPU communication bandwidth, this feature may slow down the logging process.

\textbf{Communication Optimization\hspace{2.5mm}} If training data are split across multiple processes with distributed training, we need to aggregate dataset-level statistics across processes for consistency. To minimize the communication cost, we delay the synchronization process until the training loop (one epoch) is over, and perform synchronization only once at the end. Following the similar logic, users can maximize the efficiency of the logging phase by disabling gradient synchronization (\eg,\ torch.no\_sync).
\newpage

\section{Broader Impacts \& Limitations}
\label{sec:neurips}
\subsection{Broader Impacts}
\label{sec:impact}
The data valuation problem can be a socially sensitive topic. As of now, we do not have the agreed-upon social norm for data valuation, and thus we refrained from discussing how exact data values should be determined based on our method. Rather, our work is an \textit{initial} attempt to tackle the \textit{technical} challenges in enabling LLM-scale data valuation. For equitable data valuation, we believe future research for improving both accuracy and efficiency of data valuation systems along with extensive social discussions are necessary.

\subsection{Limitations \& Future Work}
\label{sec:limitation}
We generally observed that influence function approaches are susceptible to outlier data with large gradient norms. This outlier issue is particularly severe for language modeling tasks due to the fact that the gradient of each sequence is the sum of gradients for all tokens in that sequence. If a few tokens in the sequence have large gradient norms, their gradients may dominate the total gradient for the sequence and hurt data valuation accuracy.  While our work tried to reduce the outlier effect with (self-influence) normalization, exploring other filtering heuristics (\eg,\ $L_2/L_1$ norm ratio~\cite{grosse2023studying}) may be an interesting research direction.

We attempted to lay the software foundation for data valuation with \software, but did not implement extensive system support, such as high-performance vector database (\eg,\ Faiss~\cite{johnson2019billion}). We expect further system optimizations would enable significantly more efficient data valuation. To reduce the cost of influence functions, our work mostly explored low-rank gradient projection, which compresses the gradient in a spectral domain in essence. Noting that gradient compression has been extensively studied in the efficient distributed training literature, it is worth exploring (or combining) different gradient compression strategies, \eg,\ top-$k$ compression~\cite{shi2019understanding} or low-bit compression~\cite{Wen2017TernGradTG}, to further reduce the compute/memory/storage costs for influence functions.